\documentclass[11pt,a4paper]{article}
\usepackage[hyperref]{emnlp2020}
\usepackage[utf8]{inputenc}
\usepackage{times}
\usepackage{latexsym}
\usepackage{url}
\usepackage{amssymb}
\usepackage{amsmath}
\usepackage{subfig}
\usepackage{graphicx}
\usepackage{graphics}
\usepackage{booktabs}
\usepackage{multirow}
\usepackage{url}
\usepackage{color}
\usepackage{hyperref}
\usepackage{fixltx2e}
\usepackage{enumitem}
\usepackage{pgfplots}
\usepackage{makecell}
\usepackage{comment}
\setlist[itemize,enumerate]{leftmargin=*}

\usepackage{microtype}

\newcommand{\aspas}[1]{``#1''}
\newcommand{\til}{$\sim$}

\DeclareMathOperator*{\argmax}{\text{arg\,max}}

\newcommand{\remove}[1]{}

\pgfplotsset{width=10cm,compat=1.9}
\pgfplotscreateplotcyclelist{my black white}{%
    blue!90!white, solid, every mark/.append style={solid, fill=blue!90!white}, mark=square*\\%
    red!90!white, dashed, every mark/.append style={solid, fill=red!90!white},mark=diamond*\\%
    green!40!black, densely dotted, every mark/.append style={solid, fill=green!40!black}, mark=*\\%
    loosely dashed, every mark/.append style={solid, fill=gray},mark=*\\%
    densely dashed, every mark/.append style={solid, fill=gray},mark=square*\\%
    dashdotted, every mark/.append style={solid, fill=gray},mark=otimes*\\%
    dashdotdotted, every mark/.append style={solid},mark=star\\%
    densely dashdotted,every mark/.append style={solid, fill=gray},mark=diamond*\\%
}

\aclfinalcopy % Uncomment this line for the final submission
\title{The Explanation Game:\\Towards Prediction Explainability through Sparse Communication}
\author{Marcos V. Treviso \\
  Instituto de Telecomunica\c{c}\~oes\\
  Instituto Superior Técnico\\
  University of Lisbon, Portugal\\
  \texttt{\small marcos.treviso@tecnico.ulisboa.pt} 
  \And 
  Andr\'{e} F.~T. Martins \\
  Instituto de Telecomunica\c{c}\~oes \\
  LUMLIS (Lisbon ELLIS Unit) \\
  Instituto Superior Técnico \& Unbabel \\
  Lisbon, Portugal\\
  \texttt{\small andre.t.martins@tecnico.ulisboa.com}}

\date{}

\begin{document}
\maketitle

\begin{abstract}
Explainability is a topic of growing importance in  NLP. In this work, we provide a unified perspective of explainability as a communication problem between an explainer and a layperson about a classifier's decision. 
We use this framework to compare several %prior approaches for extracting explanations  
explainers, 
including gradient methods, %representation 
erasure, and attention mechanisms, 
in terms of their communication success. 
In addition, we reinterpret these methods in the light of classical feature selection, and use this as inspiration for new  embedded explainers, through the use of selective, sparse attention. 
Experiments in text classification and natural language inference,\remove{and machine translation, } using different configurations of explainers and laypeople (including both machines and humans), reveal an advantage of attention-based explainers over gradient and erasure methods, and show that selective attention is a simpler alternative to stochastic rationalizers. Human experiments show strong results on text classification with  post-hoc explainers trained to optimize communication success.  
% \textcolor{red}{todo: include selective attention advantage over rationalizers?}
%Human experiments show promising results on text classification with  post-hoc explainers trained to optimize communication success. % and faithfulness.
\end{abstract}

\section{Introduction}

The widespread use of machine learning to assist humans in decision making brings the need for explaining models' predictions \citep{doshi2017towards,lipton2016mythos,rudin2019stop,miller2019explanation}. 
This poses a challenge in NLP, where current state-of-the-art neural systems are generally opaque 
\citep{goldberg2017neural,peters2018deep,devlin2019bert}. 
Despite the large body of recent work  (reviewed in \S\ref{sec:related_work}), a unified perspective modeling the human-machine interaction---a {\it communication} process in its essence---is still missing.

\begin{figure}[t]
    \centering
    % \andre{figure with explainer and layperson}
    % \includegraphics[width=0.8\textwidth]{figs/framework-3.pdf}
    \includegraphics[width=1\columnwidth]{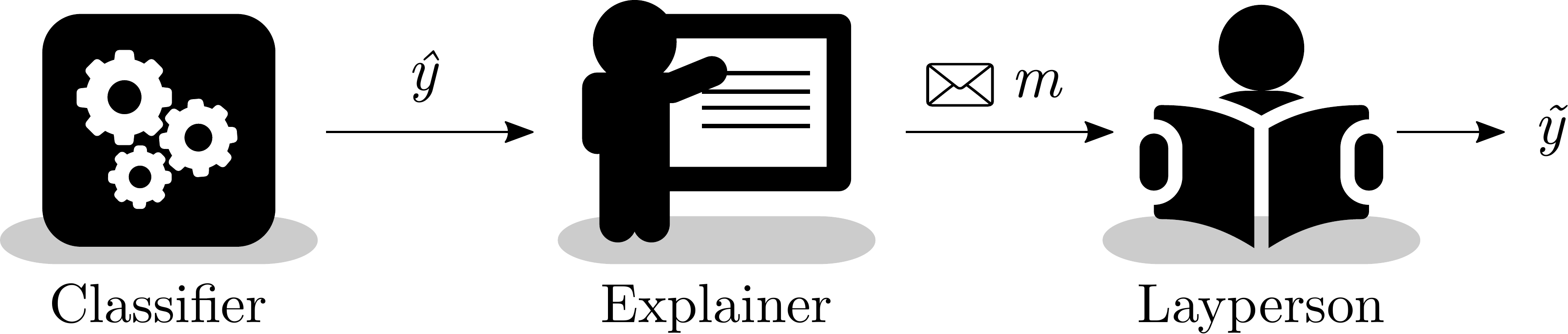}
    \caption{Our framework to model explainability as communication. Predictions $\hat{y}$ are made by a classifier $C$; an explainer $E$ (either embedded in $C$ or operating post-hoc) accesses these predictions and communicates an explanation (a message $m$) to the layperson $L$. Success of the communication is dictated by the ability of $L$ and $C$ to match their predictions: $\tilde{y} \stackrel{?}{=} \hat{y}$. Both the explainer and layperson can be humans or machines.}
    \label{fig:explainer_layperson}
\end{figure}

%Deep neural networks are used successfully in a range of NLP taks, achieving state-of-the-art results in various tasks, like sentiment classification \citep{thongtan-phienthrakul-2019-sentiment}, named entity recognition \citep{Liu2019gcdt}, summarization \citep{unilm} and machine translation \citep{edunov-etal-2018-understanding}. 
%\andre{the choice of tasks and papers to cite seems a bit arbitrary here, we can probably skip them and cite a recent book or survey, such as \citep{goldberg2017neural} or \citep{manning2015computational}} 

%, and despite recent advances in methods for extracting explanations and rationales \citep{ribeiro2016should,lei2016rationalizing}, they still fall under the categorization of black-boxes. 

Many methods have been proposed to generate explanations. 
Some neural network architectures are equipped with built-in components---{attention mechanisms}---which weigh the relevance of input features for triggering a decision  \cite{Bahdanau2015,vaswani2017attention}. Top-$k$ attention weights 
provide plausible, but not always faithful, explanations  \citep{jain2019attention,serrano2019attention,wiegreffe2019attention}. 
Rationalizers with hard attention are arguably more faithful, but require stochastic networks, which are harder to train   \citep{lei2016rationalizing,bastings2019interpretable}. 
Other approaches 
include gradient methods \cite{li2016visualizing,arras-etal-2017-explaining}, querying the classifier with leave-one-out strategies \citep{li2016visualizing,feng2018pathologies}, or training local sparse classifiers \cite{ribeiro2016should}. 

%seek local explanations by evaluating the {gradient} of the predicted label with respect to the input features \cite{li2016visualizing,arras-etal-2017-explaining}, or in a post-hoc manner by training a  sparse linear model on a vicinity of the input example \cite{ribeiro2016should} or by repeatedly querying the classifier with leave-one-out strategies \citep{li2016visualizing,feng2018pathologies}.

% How to compare these different approaches? 
How should these different approaches be compared?
Several diagnostic tests have been proposed: \citet{jain2019attention} assessed the explanatory power of attention weights by measuring their correlation with input gradients; 
\citet{wiegreffe2019attention} and \citet{deyoung2019eraser} developed more informative tests, including a combination of comprehensiveness and sufficiency metrics and the correlation with human rationales; 
\citet{jacovi2020towards} proposed a set of evaluation recommendations and a graded notion of faithfulness. Most proposed frameworks rely on correlations and counterfactual simulation, sidestepping the main practical goal of prediction explainability---the ability to {\it communicate} an explanation to a human user.

%Prior work proposed various diagnostic tests for assessing the explanatory power of the different methods, for example by measuring their correlation with input gradients \cite{jain2019attention} has diagnosed the explanatory power of attention weights by measuring their correlation with input gradients, but is this a diagnostic tool or rather a comparison between two explanatory methods? 

In this work, we fill the gap above by proposing a unified framework that regards explainability as a {\bf communication problem}. %how can a machine communicate a justification for its decision (either a faithful explanation or a post-hoc interpretation) to a human? 
Our framework is inspired by human-grounded evaluation through {\bf forward simulation/prediction}, as proposed by  \citet[\S3.2]{doshi2017towards}, where humans are presented with an explanation and an input, and
must correctly simulate the model's output (regardless of the true output). 
We model this process as shown in Figure~\ref{fig:explainer_layperson}, by considering the interaction between a {\it classifier} (the  model whose predictions we want to explain), an {\it explainer} (which provides the explanations), and a {\it layperson} (which must recover the classifier's prediction). 
We show that different configurations of these components correspond to previously proposed explanation methods, and we experiment with explainers and laypeople being both humans and machines. 
%Our framework recovers as particular cases many previously proposed explainers, 
%according to a typology that draws a connection between traditional feature selection  \citep{guyon2003introduction} and modern explanation techniques. 
%and it also inspires new ones: 
Our framework also inspires two new methods: 
%It also allows to test new variants: 
embedded explainers based on {\bf selective attention} \citep{martins2016softmax,peters2019sparse}, and {\bf trainable explainers} based on emergent communication \citep{foerster2016learning,lazaridou2016multi}. 

%\marcos{Taking into consideration the typology of plausible and faithful explanations distinguished by \citet{wiegreffe2019attention}, the goal of this paper is to provide a framework that gives a quantitative characterization for when a explanation should be considered plausible. As discussed by \citet[\S5]{wiegreffe2019attention}, plausible explanations are very important even if not faithful: \citet{rudin2019stop} defines explainability as a plausible reconstruction of the decision-making process, and \citet{riedl2019human} argues that they mimic what humans do when rationalizing past actions. We discuss how our framework is related with previous works in \S\ref{sec:related_work}.}

% we focus on better characterizing when explanations should be considered *plausible*, by providing a framework that gives an objective answer to this question. As discussed in Wiegreffe and Pinter, 2019, §5, plausible explanations are very important even if not faithful: Rudin (2018) defines explainability as a plausible reconstruction of the decision-making process, and Riedl (2019) argues that they mimic what humans do when rationalizing past actions. 

%We model explainability as communication 
%under the principle of parsimony (``Occam's razor''), where the goal is for an explainer to communicate the rationale for a model's decision to a layperson using the least possible words. 
%These explanations should be informative (about the decision itself and the model's decision process), compact, and understandable by humans.

Overall, our contributions are:
\begin{itemize}
    \item We draw a link between recent techniques for explainability of neural networks and classic feature selection in linear models   (\S\ref{sec:feature_selection}). This leads to 
    %\item We propose 
    new embedded methods for explainability through selective, sparse attention (\S\ref{sec:attention}).
    \item We propose a new framework to assess explanatory power as the communication success rate between an explainer and a layperson %The success of communication serves as an assessment of explanation quality
    (\S\ref{sec:communication}). 
    \item We experiment with text classification, natural language inference, and machine translation, using different configurations of explainers and laypeople, both machines (\S\ref{sec:experiments}) and humans (\S\ref{sec:human_evaluation}). %. This allows comparing in the same grounds models that use attention-based interpretations, post-hoc explanations, and gradient-based information 
\end{itemize}

\section{Revisiting Feature Selection}\label{sec:feature_selection}

A common way of generating explanations is by highlighting {\it rationales} \citep{zaidan2008modeling}.  %\citep{lei2016rationalizing,ribeiro2016should}. 
%A motivation for keeping this set small is the 
The principle of parsimony (``Occam's razor'') advocates %according to which 
simple explanations %should be preferred 
over complex ones. 
This principle  inspired a large body of work in traditional feature selection for linear models.  We draw here a link between that work and modern approaches to explainability. 

%We start by drawing a link between explainability of neural networks and feature selection in linear models, making a bridge between the two worlds. 
%This connection is tied to the distinction between model interpretability and prediction explanations made by \citet{lipton2016mythos}.

%\andre{perhaps bring some notation here, where $\boldsymbol{W}$ are the weights of a linear model, $\boldsymbol{\phi}(\boldsymbol{x})$ are the features for an input $\boldsymbol{x}$, many of which are believed to be irrelevant for any input---if the $j$th feature $\phi_j$ is irrelevant, setting the $j$th column of $\boldsymbol{W}$ to zero eliminates that feature from the model. 
%Sometimes a measure of feature importance is the magnitude $|w_{kj}|$, which since the model is linear happens to equal the gradient $\frac{\partial f_k(x)}{\partial \phi_j(x)}$.}

\begin{table*}[t]
    \centering
    \small
    \begin{tabular}{lp{0.43\textwidth}p{0.43\textwidth}}
    \toprule
        & {\bf Static selection} (model interpretability) & {\bf Dynamic selection} (prediction explainability) \\
    \midrule
        {\bf Wrappers} & Forward selection, backward elimination \citep{kohavi1997wrappers} & Input reduction \citep{feng2018pathologies}, representation erasure (leave-one-out) \citep{li2016understanding,serrano2019attention}, LIME \citep{ribeiro2016should}\\
    \midrule
        {\bf Filters} & Pointwise mutual information \citep{church1990word}, %feature counts, 
        recursive feature elimination \citep{guyon2002gene} & Input gradient \citep{li2016visualizing}, layerwise relevance propagation \citep{bach2015pixel}, top-$k$ softmax attention \\
    \midrule
       {\bf Embedded} & $\ell_1$-regularization \citep{tibshirani1996regression}, elastic net \citep{zou2005regularization} &  Stochastic attention \citep{xu2015show,lei2016rationalizing,bastings2019interpretable}, sparse attention ({\bf this paper}, \S\ref{sec:attention})  \\
    
    \bottomrule
    \end{tabular}
    \caption{Overview of static and dynamic feature selection techniques.} %for model interpretability and prediction explainability.
    \label{tab:feature_selection}
\end{table*}

Table~\ref{tab:feature_selection} highlights the connections. 
Traditional feature selection methods \citep{guyon2003introduction} 
are mostly concerned with {\bf model interpretability}, i.e., understanding how  models behave globally. 
%\toremove{There is a rich body of literature applying these methods to linear feature-based models, with the goal of making sense of a large number of automatically extracted features, many of which are irrelevant for the task at hand.}
Feature selection happens {\it statically} during model training, after which irrelevant features are permanently deleted from the model. 
This contrasts with {\bf prediction explainability} in neural networks, 
where feature selection happens {\it dynamically} at run time: 
%\toremove{Given a particular input and the corresponding model's decision, the goal is to provide a local explanation, understandable by humans, that justifies the model's decision}.
here explanations are input-dependent, hence a feature not relevant for a particular input can be relevant for another. 
%Commonly used explanations are highlighted input features (such as words in a document) or prototypes (similar examples in the training data). 
%This paper focuses on the first case, but the framework we propose here is applicable to the second case as well. 
Are these two worlds far away? 
\citet[\S 4]{guyon2003introduction} proposed a typology for traditional feature selection 
with three classes of methods, 
distinguished by how they model the interaction between their main two components, 
the {\it feature selector} and the {\it learning algorithm}. We argue that this typology can also be used to characterize various explanation methods, if we replace these two components by the {\it explainer} $E$ and the {\it classifier} $C$, respectively. 
\begin{itemize}
\item {\bf Wrapper methods}, in the wording of \citet{guyon2003introduction}, 
``utilize the learning machine of interest as a black box to score subsets of
variables according to their predictive power.'' 
This means greedily searching over subsets of features, training a model with each candidate subset. 
In the dynamic feature selection world, 
this is somewhat reminiscent of the leave-one-out method of \citet{li2016understanding}, the ablative approach of \citet{serrano2019attention}, and LIME \citep{ribeiro2016should}, which repeatedly queries the classifier to label new examples. 
%This class of methods requires  multiple calls to the learning algorithm (in the case of static feature selection) or the classifier $C$ (in the case of dynamic feature selection).
\item {\bf Filter methods} decide to include/exclude a feature based on an  importance metric (such as feature counts or pairwise mutual information). This can be done as a preprocessing step or by training the model once and thresholding the feature weights. %; the latter can be done in combination with a wrapper, as in the recursive feature elimination algorithm \citep{guyon2002gene}. 
In dynamic feature selection, this is done when we examine the gradient of the prediction with respect to each input feature, and then select the features whose gradients have large magnitude \citep{li2016visualizing,arras2016explaining,jain2019attention},%
\footnote{In linear models this gradient equals the feature's weight.} %
%This is the approach taken by the input gradient and layerwise relevance propagation techniques 
%\citep{li2016visualizing,arras2016explaining,jain2019attention}, where this gradient is taken as a measure of feature ``importance.'' 
and when thresholding softmax attention scores to select relevant input features, as analyzed by \citet{jain2019attention} and \citet{wiegreffe2019attention}. 
%This class of methods involve at most one call to the classifier $C$. 
%\andre{may say after the bullet points that this sort of attention is not really an embedded method.}
\item {\bf Embedded methods}, in traditional feature selection, embed feature selection within the learning algorithm by using a sparse regularizer such as the $\ell_1$-norm  \citep{tibshirani1996regression}.
Features that receive zero weight become irrelevant and can be removed from the model. 
In dynamic feature selection, this encompasses methods where 
the classifier produces rationales together with its decisions \citep{lei2016rationalizing,bastings2019interpretable}. 
We propose in \S\ref{sec:attention} an alternative approach via  {\bf sparse attention} \citep{martins2016softmax,peters2019sparse}, where 
the selection of words for the rationale resembles  $\ell_1$-regularization. 

\end{itemize}

In \S\ref{sec:communication}, we  frame each of the cases above as a communication process, where the explainer $E$ aims to communicate a short message with the relevant features that triggered the classifier $C$'s decisions to a layperson $L$. The three cases above are distinguished by the way $C$ and $E$ interact.

\section{Embedded Sparse Attention}\label{sec:attention}

%\andre{this needs to be completed. may be a full section or a subsection of the previous one.}

The case where the explainer $E$ is embedded in the classifier $C$ naturally favors faithfulness, since the mechanism that explains the decision (the {\it why}) can also influence it (the {\it how}). 

{Attention mechanisms} \citep{Bahdanau2015} allow visualizing relevant input features that contributed to the model's decision. 
%Attention weights are usually computed by applying a softmax transformation to a vector of scores $\mathbf{s} \in \mathbb{R}^n$, which are themselves a function of a query and key vectors. 
However, the traditional softmax-based attention is {\it dense}, i.e., it gives {\it some} probability mass to every feature, even if small. 
The typical approach is to select the top-$k$ words with largest attention weights as the explanation. However, this is not a truly embedded method, but rather a filter, and as pointed out by \citet{jain2019attention} and \citet{wiegreffe2019attention}, it may not lead to faithful explanations. 
%In the world of static feature selection, this is similar to what happens when using  $\ell_2$-regularization for filtering: features whose weights have  small magnitude are good candidates to remove from the model, i.e., their weights can be truncated to zero. 

An alternative is to embed in the classifier an attention mechanism that is inherently {\bf selective}, i.e., which can produce sparse attention distributions natively, where some input features receive exactly zero attention. 
An extreme example is hard attention, which, as argued by \citet{deyoung2019eraser}, provides more faithful explanations ``by construction'' as they discretely extract snippets from the input to pass to the
classifier. 
A problem with hard attention is its non-differentiability, which complicates training \citep{lei2016rationalizing,bastings2019interpretable}. 
We consider in this paper a different approach: using end-to-end differentiable sparse attention mechanisms, via the {\bf sparsemax} \citep{martins2016softmax} and the recently proposed {\bf 1.5-entmax} transformation \citep{peters2019sparse}, described in detail in \S\ref{sec:supp_attention}. 
These sparse attention transformations have been applied successfully to machine translation and morphological inflection  \citep{peters2019sparse,correia2019adaptively}. 
%By using sparsemax or 1.5-entmax attention, we obtain a selective attention mechanism: 
Words that receive non-zero attention probability are {\it selected} to be part of the explanation. This is an embedded method akin of the use of $\ell_1$-regularization in static feature selection.  
We experiment with these sparse attention mechanisms in \S\ref{sec:experiments}.

\section{Explainability as Communication}\label{sec:communication}

We now have the necessary ingredients to describe our unified framework for comparing and designing explanation strategies, illustrated in Figure~\ref{fig:explainer_layperson}. 

Our fundamental assumption is that explainability is intimately linked to the ability of an explainer to {\bf communicate} the rationale of a decision in terms that can be understood by a human; 
we use the success of this communication as a criterion for how plausible the explanation is.  

\subsection{The~Classifier-Explainer-Layperson~setup} \label{subsec:setup_clf_exp_lay}

Our framework draws inspiration from Lewis' signaling games \citep{lewis1969convention} and the recent work on emergent communication  \citep{foerster2016learning,lazaridou2016multi,havrylov2017emergence}. 
% \marcos{Are these the correct references pointed by the third reviewer?}
% \marcos{yes.} He mentioned (Foerster, 2016) and (Lazaridou et al)
Our starting point is the classifier $C: \mathcal{X} \rightarrow \mathcal{Y}$ which, when given an input $x \in \mathcal{X}$, produces a prediction $\hat{y} \in \mathcal{Y}$. 
This is the prediction that we want to explain. 
An explanation is a {\bf message} $m \in \mathcal{M}$, for a predefined message space $\mathcal{M}$ (for example, a rationale). 
%We say that the predictions made by $C$ are {\it explainable} if there is an explainer $E$ that can
The goal of the explainer $E$ is to compose and {\bf successfully communicate} messages $m$ to a layperson $L$. The success of the communication is dictated by the ability of $L$ to reconstruct $\hat{y}$ from $m$ with high accuracy. 
In this paper, we experiment with $E$ and $L$ being either humans or machines. 
Our framework is inspired by human-grounded evaluation through forward simulation/prediction, as proposed by \citet[\S3.2]{doshi2017towards}. 
More formally:
\begin{itemize}
    \item The \textbf{classifier} $C$ is the model whose predictions we want to explain. For given inputs $x$,
    % (in this paper, documents or sentences)
    $C$ produces $\hat{y}$ that are hopefully close to the ground truth $y$. We are agnostic about the kind of model used as a classifier, but we assume that it computes certain internal representations $h$ that can be exposed to the explainer. %\toremove{For instance, $h$ could be internal representations in a neural network.} %\andre{check if we need this}
    
    \item The \textbf{explainer} $E$ produces explanations for $C$'s decisions. It receives the input $x$, the classifier prediction $\hat{y} = C(x)$, and optionally the internal representations $h$ exposed by $C$. It outputs a message $m \in \mathcal{M}$ regarded as a \aspas{rationale} for $\hat{y}$. The message $m = E(x, \hat{y}, h)$ should be simple and compact enough to be easily transmitted and understood by the layperson $L$. 
    %A good explanation must be informative (about the model's decision and the decision process), compact, and understandable by humans. 
In this paper, %to ensure the last two points, 
we constrain  messages 
%The message space $\mathcal{M}$ must be composed of representations that are readable to humans \citep{wiegreffe2019attention}. For reasons that will be clear later, an explainer might not use all of its inputs to produce explanations, characterizing explanations with limited information.
% In this paper, %all messages are bags of words extracted from textual input. 
to be bags-of-words (BoWs) extracted from the textual input $x$.
% \footnote{Note that our framework is  flexible about the choice of this message space $\mathcal{M}$. For example, explanations could also be {\it prototypes}, i.e., small subsets of training examples.}
% , up to a maximum length of $k$ words. 
%\footnote{Note that our framework is  flexible about the choice of this message space $\mathcal{M}$. For example, explanations could also be {\it prototypes}, %and {\bf criticisms} \citep{kim2016examples}, 
%i.e., small subsets of training examples. 
%\andre{maybe mention other possibilities as well?}
%\marcos{counterfactuals? interpretable embeddings?}
%in which case $\mathcal{M}$ would consist of all possible subsets 
%(with bounded cardinality) of training examples. %\toremove{In this case, a large attention mechanism (better called a memory network in this context, \citet{weston2014memory}) could also be used to point to the relevant examples that should be part of the explanation, based on similarity with the current input.}
%} 
%\marcos{Moreover, the choice of $\mathcal{M}$ is independent of $E$, and therefore it does not favor any specific method.} \marcos{I think we need to explain the reason here.}

    % \marcos{Para machine translation a mensagem eh uma janela de embeddings. Falar disso aqui ou explicamos na subsecao de machine translation?}
    
    \item The \textbf{layperson} $L$ is a simple model (e.g., a linear classifier)%
    \footnote{The reason why we assume the layperson is a simple model is to encourage the explainer to produce simple and explanatory messages, in the sense that a simple model can learn with them. A more powerful layperson could potentially do well even with bad explanations.} %
that receives the message $m$ as input, and predicts a final output $\tilde{y} = L(m)$. The communication is successful if $\tilde{y} = \hat{y}$.  Given a test set $\{x_1, \ldots, x_N\}$, we evaluate the {\bf communication success rate} (CSR) as the fraction of examples for which the communication is successful: \begin{equation}\label{eq:comm_accuracy}
        \mathrm{CSR} = \frac{1}{N}\sum_{n=1}^N \big[\big[{C(x_n) = L(E(x_n, C(x_n)))}\big]\big],
    \end{equation}
where $[[\cdot]]$ is the Iverson bracket notation. 
\end{itemize}

Under this framework, we regard the communication success rate as a quantifiable measure of explainability: a high CSR means that the layperson $L$ is able to replicate the classifier $C$'s decisions a large fraction of the time when presented with the messages given by the explainer $E$; this assesses how informative $E$'s messages are. % for the two agents to communicate successfully.

% \marcos{I'm thinking that Cacc will make things a bit confusing - it reminds me of classifier acc} \andre{I agree and had the same thought. Also, I need to say somehwere that we omit the $h$ argument from $E$ to avoid clutter.}
% \marcos{Replaced by communication success rate CSR}

%Our framework aims to capture the fact that 
%a good explanation must be informative (about the model's decision and the decision process), compact, and understandable by humans. 

%\paragraph{Message space $\mathcal{M}$.} 
%A good explanation must be informative (about the model's decision and the decision process), compact, and understandable by humans. 
%In this paper, to ensure the last two points, we constrain the messages in $\mathcal{M}$
%to be bags of words extracted from the input document/sentence $x$, up to a maximum length of $k$ words. 
%The communication accuracy defined above can be regarded as an automatic metric of explainability. 

Our framework is flexible, allowing different configurations for  $C$, $E$, and $L$, as next described. 
In \S\ref{sec:experiments}, we show  examples of explainers and laypeople for text classification and natural language inference tasks (additional experiments on machine translation are described in \S\ref{sec:nmt_setup}).

\paragraph{Relation to filters and wrappers.} 

In the wrapper and filter approaches described in \S\ref{sec:feature_selection}, the classifier $C$ and the explainer $E$ are separate components. In these approaches, $E$ works as a {\it post-hoc explainer}, querying $C$ with new examples or  requesting gradient information. 

\paragraph{Relation to embedded explanation.} 

By contrast, in the embedded approaches of \citet{lei2016rationalizing} and the selective sparse attention introduced in \S\ref{sec:attention}, the explainer $E$ is directly {\it embedded} as an internal component of the classifier $C$, returning the selected features as the message. 
This approach is arguably more faithful, as $E$ is directly linked to the mechanism that produces $C$'s decisions.

%\paragraph{Joint training of explainer and layperson.} 

%\paragraph{Prototypes as message.} 
%\andre{maybe delete this paragraph or move it to a footnote.} 
%We can also assume that explanations are simply \textbf{pointers to examples} $\{x_i, y_i\}$ in the training set. This can also be formulated with an attention-based model, but now the attention is over the training data and not over the document words. In this case, $E$ could provide an explanation for $C$ by fetching a small set of nearest neighbors in the training data, rendering $m$ as a \aspas{bag-of-examples}.  The layperson needs only to examine the bag-of-examples $m$ and use e.g. majority voting to pick an example $x_i$ and the corresponding label $y_i$ as the final prediction $\tilde{y}$.

\subsection{Joint training of explainer and layperson}
\label{subsec:joint_e_and_l}

So far we have assumed that $E$ is given beforehand, chosen among existing explanation methods, and that $L$ is trained to assess the explanatory ability of $E$. 
But can our framework be used to {\it create} new explainers by training $E$ and $L$ jointly? 
%For communication to succeed, $E$ and $L$ have to agree on a protocol that ensures informative messages.  
We will  see how this can be done by letting $E$ and $L$ play a cooperative game \citep{lewis1969convention}. 
The key idea is that they need to learn a communication protocol that ensures high CSR (Eq.~\ref{eq:comm_accuracy}). 
Special care needs to be taken to rule out ``trivial'' protocols and ensure plausible, potentially faithful, explanations.
We propose a strategy to ensure this, which will be validated using human evaluation in \S\ref{sec:human_evaluation}.%
\footnote{Other approaches, such as \citet{lei2016rationalizing} and \citet{yu2019rethinking}, develop rationalizers from cooperative or adversarial games between generators and encoders. However, those  frameworks do not aim at explaining an external classifier.} % 

%We assume in this setting that explanations are extracted after $C$ is trained, i.e., $E$ is a post-hoc explainer. 

% \andre{Mention we're learning a post-hoc explanation model. }

%A good explanation must be {\it informative} (both about the decision and the model's decision process), {\it compact}, and {\it understandable}. 
%The last two requirements are ensured by defining  $\mathcal{M}$ properly. %in this paper, we do this by letting messages be small BoWs and limiting the maximum length of a message. 
%But how to ensure the first requirement? 
%To fully model the communication process and maximize the informativeness of the explanations $m$, we consider a strategy where both the explainer 
Let 
$E_{\theta}$ and layperson $L_{\phi}$ be {\bf trained models} (with parameters $\theta$ and $\phi$), learned together to optimize a multi-task objective  with two terms:
\begin{itemize}
    \item A {\bf reconstruction term} that controls the information about the classifier's decision $\hat{y}$. We use a  cross-entropy loss on the output of the layperson $L$, using $\hat{y}$ (and not the true label $y$) as the ground truth: $\mathcal{L}(\phi, \theta) = -\log p_\phi(\hat{y} \mid m)$, where $m$ is the output of the explainer $E_{\theta}$.
    \item A {\bf faithfulness term} that encourages  the explainer $E$ to take into account the classifier's decision process when producing its explanation $m$. This is done by adding a squared loss term $\Omega(\theta) = \|\tilde{h}(E_{\theta}), h\|^2$ where $\tilde{h}$ is $E$'s prediction of $C$'s  internal representation $h$. 
    % $\Omega(\theta) = 1-\cos(\tilde{h}(E_{\theta}), h)$. 
    %$\Omega(\theta) = \mathrm{MSE}(\tilde{h}(E_{\theta}), h)$. 
    %\marcos{say somewhere that $h$ and $\tilde{h}$ are the average of LSTM vectors. i.e. we are approximating the average of the LSTMs of $C$ and $L$}
    %This encourages the explanations to be faithful.
\end{itemize}
The objective function 
% to minimize 
is a combination of these two terms, $\mathcal{L}_{\Omega}(\phi, \theta) := \lambda \Omega(\theta) +  \mathcal{L}(\phi, \theta)$. 
We used $\lambda = 1$ in our experiments. 
This objective is minimized in a training set that contains pairs $(x,\hat{y})$. 
%\andre{Marcos, which dataset did we use? I am thinking now that if we use the same training set used to train the classifier $C$, the classifier may be over-confident in his predictions (due to over-fitting) and it will output predictions $\hat{y}$ that will be very similar to the gold $y$...} \marcos{IMDB and SNLI. This actually happens with batota maior que 0, for sparsemax it becomes even more clear that this process of cheating happens, though for imdb the explanations look very good.} 
Therefore, in this model the message $m$ is latent and works as a ``bottleneck'' for the layperson $L$, which does not have access to the full input $x$, to guess the classifier's prediction $\hat{y}$---related models have been devised in the context of emergent communication \citep{lazaridou2016multi,foerster2016learning,havrylov2017emergence} 
and sparse autoencoders \cite{trifonov2018learning,subramanian2018spine}. 

We minimize the objective above with gradient backpropagation. To ensure end-to-end differentiability, during this joint training we use sparsemax attention (\S\ref{sec:attention}) to select the relevant words in the message. 
% At test time, if this results in more than $k$ words, we cap the message to the $k$ most attended words only. 
One important concern in this model is to prevent $E$ and $L$ from learning a trivial protocol to maximize CSR. 
% To prevent this,
% from happening, %we need to make sure the explainer is forced to produce informative explanations that are not just
%a trivial re-encoding of the label. 
%To ensure this, %we force the explainer, for a fraction of the examples (\andre{how many?}) to generate messages $m$ based on the input $x$ only, i.e., without access to the classifier's prediction $\hat{y}$. 
%More formally, 
% we let the explainer access $\hat{y}$
% the predictions of the classifier ($\hat{y}$) 
% during training by following a linear increasing schedule, so that at the end of training the explainer has a probability of 20\% of accessing $\hat{y}$
% the explainer is only exposed to $\hat{y}$ 
% during training by following a linear increasing schedule, so that at the end of training the explainer has a probability of $\beta$ (e.g. 50\%) of accessing $\hat{y}$.
% : $\beta \cdot i / N$,  where $i$ is the iteration counter.
% with a probability of $\beta \cdot i / N $, where $i$ is the iteration counter, $N$ is the number of training examples and $\beta$ is a predefined hyperparameter (e.g. 20\%). %\marcos{Andre, can you check this?}
To ensure this, we forbid $E$ from including stopwords in its messages 
% \footnote{We used a list of 127 English stopwords from NLTK.} %
and during training we use a linear schedule for the probability of the explainer accessing the predictions of the classifier ($\hat{y}$), which are hidden otherwise. At the end of training, the explainer will access it with probability $\beta$. 
In our experiments, we set $\beta$ to 20\% (chosen on the validation set as described in \S\ref{sec:analysis_beta}).

\section{Experiments}\label{sec:experiments}

We experimented with our framework on two NLP tasks: text classification and natural language inference. Additional experiments on machine translation are reported in \S\ref{sec:nmt_setup}, with similar conclusions.

\remove{
\begin{figure}[t]
    \centering
    \includegraphics[width=1\columnwidth]{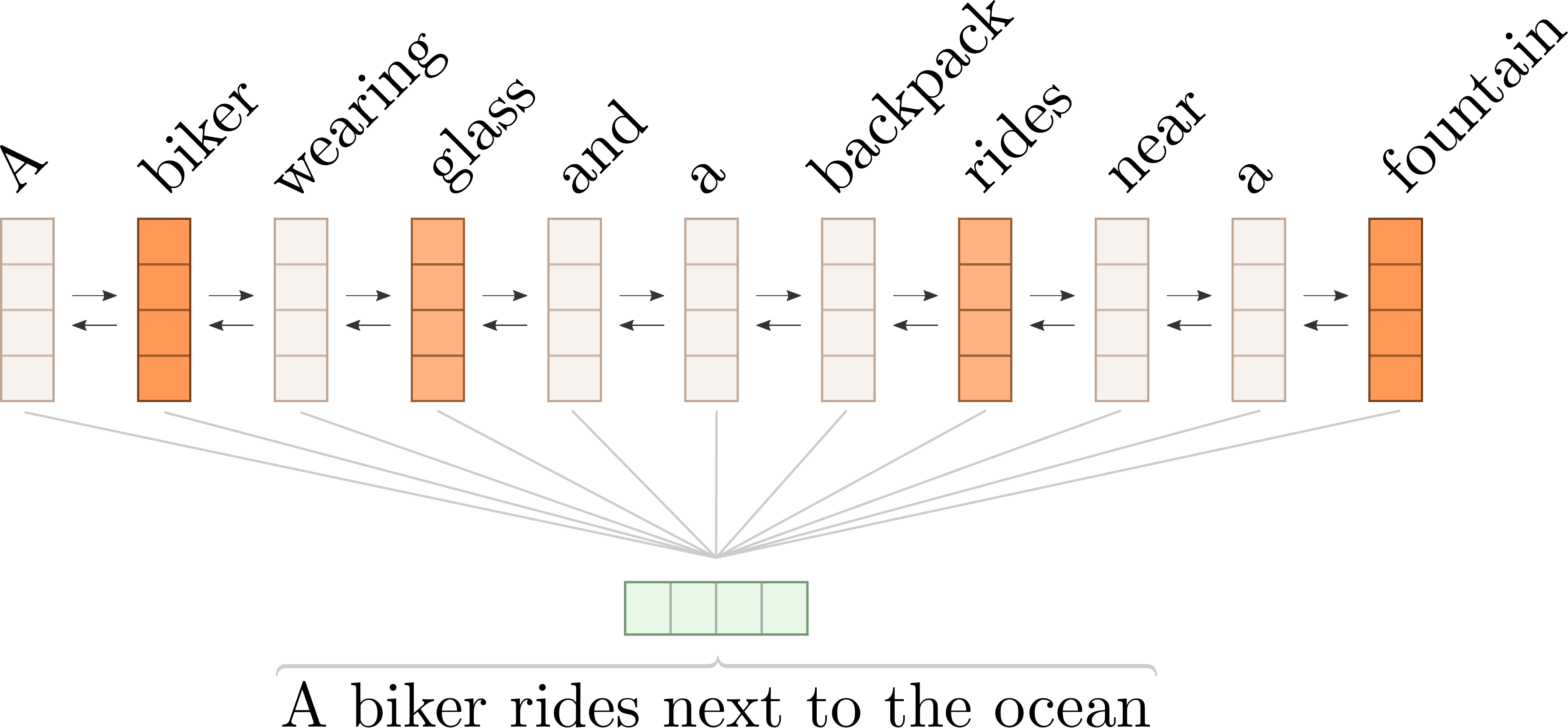}
    \caption{\toremove{Example of sparse attention for natural language inference. The selected premise words (``biker'', ``glass'', ``rides'', and ``fountain'') form the message, together with the hypothesis.}}
    \label{fig:example_nli}
\end{figure}
}

%
%\footnote{The code and human annotations will be released upon acceptance. \andre{TODO: add this for the EMNLP version}} 

%We compared different methods from the typology described in \S\ref{sec:feature_selection} and we conducted experiments using both machines and humans as explainers and laypeople.

%\subsection{Text classification and NLI}\label{sec:experiments_class}

%\paragraph{Datasets.} 
We used 4 datasets (SST, IMDB, AgNews, Yelp) for text classification and one dataset (SNLI) for NLI, with statistics  and details in Table~\ref{table:datasets} (\S\ref{sec:data_statistics_preparations}).  

\paragraph{Classifier $C$.} 
For text classification, the input $x \in \mathcal{X}$ is a document and the output set $\mathcal{Y}$ is a set of labels (e.g. topics or sentiment labels). The message is a bag of words (BoW) extracted from the document. 
As in \citet{jain2019attention} and \citet{wiegreffe2019attention}, our classifier $C$ is an RNN with attention. 
%% For machine translation, we use the encoded vector of the word being translated as the query and all the source encoded vectors as keys.  \marcos{vou explicar mais a frente isso, aqui so vai deixar mais confuso.}
%We used this BiLSTM with attention as the classifier $C$. 
For NLI, the input $x$ is a pair of sentences (premise and hypothesis) and the labels in $\mathcal{Y}$ are entailment, contradiction, and neutral. 
We let messages be again BoWs, and we constrain them to be selected from the premise (and concatenated with the full hypothesis). 
We used a similar classifier as above, but with two independent BiLSTM layers, one for each sentence. We  used the additive attention of \citet{Bahdanau2015} with the last hidden state of the hypothesis as the query and the premise vectors as keys. 

We also experimented with RNN classifiers that replace softmax attention by 1.5-entmax ($C_{\mathrm{ent}}$) and sparsemax ($C_{\mathrm{sp}}$), and with the rationalizer models of \citet{lei2016rationalizing} ($C_\mathrm{bern}$) and \citet{bastings2019interpretable} ($C_\mathrm{hk}$). 
Details about these classifiers and their hyperparameters are listed in \S\ref{sec:training_details}.  
%We use the available implementation of both models provided by the latter.%
% \footnote{We use the implementation in \url{https://github.com/bastings/interpretable_predictions}}
Table~\ref{table:results_classifiers} reports the accuracy of all classifiers used in our experiments. 
The attention-based models all perform very similarly and generally better than the rationalizer models, except for SNLI, where the latter use a stronger model with decomposable attention. As expected, in general, all these classifiers outperform a bag-of-words model which is the model we use as the layperson.

\begin{table}[t]
\small

\begin{center}
% \resizebox{\textwidth}{!}{%
\begin{tabular}{@{\hspace{2.5pt}}l@{\hspace{4.5pt}}p{.12\columnwidth}@{\hspace{4.5pt}}p{.12\columnwidth}@{\hspace{4.5pt}}p{.12\columnwidth}@{\hspace{4.5pt}}p{.12\columnwidth}@{\hspace{5pt}}p{.11\columnwidth}@{\hspace{0pt}}}

\toprule
{\sc Classifier} & 
{\sc SST} & {\sc IMDB} & {\sc AgN.} & {\sc Yelp} &{\sc SNLI} \\

\midrule 

{BoW ($L$)}   	&  {82.54}   	& {88.96}  	& {95.62}   	& {68.78}   	&  {69.81}  \\
{RNN, softmax ($C$)}	         & {86.16}   	 & {\bf 91.79}   & {96.28}   	 & {\bf 75.80}    & {78.34}  \\
{--,1.5-entmax ($C_{\text{ent}}$)}		& {86.11}   	 & {91.72}  	 & {96.30}   	 & {75.72}   	 & {79.20}  \\
{--, sparsemax	($C_{\text{sp}}$)} 	 & {\bf 86.27}    & {91.52}   & {96.37}   	 & {75.72}   	 & {78.78}  \\
% {Bernoulli ($C_{\text{bern}}$)}		     & {81.99}   	 & {87.65}  	 & {95.68}    & {75.06}   	 & {85.69*}  \\
% {Bernoulli ($C_{\text{bern}}$)}		     & {81.99}   	 & {87.65}  	 & {95.68}    & {70.12}   	 & {79.24}  \\
% {HardKuma ($C_{\text{hk}}$)}	    	 & {84.13}    & {90.52}   & {\bf 96.38}   	 & {74.36}   	& {\bf 85.49}  \\
{Bernoulli ($C_{\text{bern}}$)}		     & {81.99}    & {86.99}  	 & {95.68}    & {70.12}   	 & {79.24}  \\
{HardKuma ($C_{\text{hk}}$)}	    	 & {84.13}    & {91.06}   & {\bf 96.38}   	 & {74.36}   	& {\bf 85.49}  \\

\bottomrule
\end{tabular}
% }
\end{center}
\caption{Accuracies of the original classifiers on text classification and natural language inference.} 
\label{table:results_classifiers}

\end{table}

\begin{table*}[!htb]
\small

\begin{center}
% \resizebox{\textwidth}{!}{%
\begin{tabular}{l@{  }l cc@{ }c cc@{ }c cc@{ }c cc@{ }c cc}
\toprule
& & 
\multicolumn{2}{c}{\sc SST} & & \multicolumn{2}{c}{\sc IMDB} & & \multicolumn{2}{c}{\sc AgNews} & & \multicolumn{2}{c}{\sc Yelp} & & \multicolumn{2}{c}{\sc SNLI} \\

\cmidrule{3-4} \cmidrule{6-7} \cmidrule{9-10} \cmidrule{12-13}  \cmidrule{15-16} 

\sc Clf.\,\, & 
\multirow{1}{*}{\sc Explainer} & CSR & ACC$_L$ && CSR & ACC$_L$ && CSR & ACC$_L$ && CSR & ACC$_L$ && CSR & ACC$_L$ \\

\midrule 

%Top-$k$ Uniform	 		        & 68.31   & 68.86   	& & 66.77   & 66.47   	& & 92.44   & 91.26   	& & 58.27   & 53.03   	& & 75.44   & 68.24  \\
$C$ & Random	        & 69.41   & 70.07   	& & 67.30   & 66.67   	& & 92.38   & 91.14   	& & 58.27   & 53.06   	& & 75.83   & 68.74  \\
% Top-$k$ max-out softmax	\andre{remove?}        & 75.34   & 75.23   	    & & 89.68   & 87.34   	    & & 96.33   & 94.79   	    & & 77.43   & 67.92   	& & 77.30   & 70.04  \\

$C$ & Erasure   	    & 80.12   & 81.22   	    & & 92.17   & 88.72  	& & 97.31   & 95.41   	& & 78.72   & 68.90   	& & 77.88   & 70.04  \\
% Erasure 1.5-entmax \andre{remove?}  	& 84.51   & 83.03   	    & & 92.54   & 89.06  	& & 97.21   & 95.62   	& & 80.14   & 69.67   	& & 79.27   & 72.68  \\
% Erasure sparsemax \andre{remove?} 	& \bf 85.23   & 81.82   	& & 92.87   & 89.18  	& & 95.96   & 94.51   	& & 80.53   & 69.89   	& & 82.20   & 73.86  \\

%\midrule

$C$ & Top-$k$ gradient       	& 79.35   & 79.24   	    & & 86.30   & 83.93  	    & & 96.49   & 94.86   	    & & 70.54   & 62.86   	& & 76.74   & 69.40  \\
$C$ & Top-$k$ softmax	 	    & 84.18   & 82.43   	    & & 93.06   & 89.46   	    & & \bf 97.59   & 95.61   	& & 81.00   & 70.18   	& & 78.66   & 71.00  \\
$C_{\text{ent}}$ & Top-$k$ 1.5-entmax		& \bf 85.23   & \bf 83.31   & & 93.32   & 89.60   	    & & 97.29   & \bf 95.67   	& & 82.20   & 70.78   	& & 80.23   & 73.39  \\
$C_{\text{sp}}$ & Top-$k$ sparsemax	 	& \bf 85.23   & 81.93   	& & 93.34   & 89.57   	& & 95.92   & 94.48   	    & & 82.50   & 70.99   	& & \bf 82.89   & \bf 74.76  \\

\midrule

$C_{\text{ent}}$ & Selec. 1.5-entmax		& 83.96   & 82.15   	    & & 92.55   & 89.96   	& & 97.30   & 95.66   	& & 81.38   & 70.41   	        & & 77.25   & 71.44  \\
$C_{\text{sp}}$ & Selec. sparsemax	 		& \bf 85.23   & 81.93   	& & 93.24   & 89.66   	    & & 95.92   & 94.48   	& & 83.55   & 71.60   	        & & 82.04   & 73.46  \\
% $C_{\text{bern}}$ & Bernoulli	         	& 82.37   & 78.42   	    & & 90.27   & 86.45   	    & & 96.91   & 94.43   	& & 81.21   & 69.30   	        & & 72.46*   & 68.99*  \\

% $C_{\text{bern}}$  & Bernoulli	         	& 82.37   & 78.42   	    & & 90.27   & 86.45   	    & & 96.91   & 94.43   	& & 84.93   & 66.89   	        & & 76.81   & 69.65  \\
% $C_{\text{hk}}$ & HardKuma	 		        & 85.17   & 80.40   	    & & 91.81   & 89.36   	    & & 97.11   & 95.45   	& & \bf 87.39   & \bf 71.64   	& & 74.98   & 71.48  \\

$C_{\text{bern}}$  & Bernoulli	         	& 82.37   & 78.42   	    & & 91.66   & 86.13   	    & & 96.91   & 94.43   	& & 84.93   & 66.89   	        & & 76.81   & 69.65  \\
$C_{\text{hk}}$ & HardKuma	 		        & 85.17   & 80.40   	    & & \bf 94.72   & \bf 90.16   	    & & 97.11   & 95.45   	& & \bf 87.39   & \bf 71.64   	& & 74.98   & 71.48  \\

% lambda0 is for bernoullis and selection for hardkumas
% SST: bernoulli_sparsity01     -> lambda0 = 0.01   and selection = 0.3
% IMDB: bernoulli_sparsity0003  -> lambda0 = 0.0003 and selection = 0.3
% AGNEWS: bernoulli_sparsity01  -> lambda0 = 0.01   and selection = 0.3
% YELP: bernoulli_sparsity01/   -> lambda0 = 0.01   and selection = 0.3
% SNLI: bernoulli_sparsity0003_fix_new_indep_qk/    -> lambda 0 = 0.0003 and selection = 0.1
% Hardkuma: 

% SNLI: bernoulli_sparsity001_fix_new_indep_qk/
% k: 14.72
% acc c: 80.00
% csr: 71.08
% acc l: 62.87

% SNLI: bernoulli_sparsity0003_fix_new_indep_qk/   (running again - acc l seems very low)
% k: 15.04
% acc c: 79.24
% csr: 75.37 | 76.81
% acc l: 68.62 | 69.65

\bottomrule
\end{tabular}
% }
\end{center}
\caption{CSR and layperson accuracy (ACC$_L$) for several explainers. For each explainer, we indicate the corresponding classifier from Table~\ref{table:results_classifiers}; in all cases the layperson is a BoW model. 
Only explainers of the same classifier can be compared in terms of CSR.
Top rows report performance for random, wrapper and filter explainers, for fixed $k$-word messages (the values of $k$ for the several datasets are \{5, 10, 10, 10, 4\}, respectively). Bottom rows correspond to embedded methods where $k$ is given automatically via sparsity.
The average $k$ obtained by 1.5-entmax, sparsemax, Bernoulli and HardKuma are: 
SST: \{4.65, 2.59, 6.10, 4.82\}; 
% IMDB: \{28.23, 12.94, 61.46, 74.29\}; 
IMDB: \{28.23, 12.94, 39.40, 24.18\}; 
\textsc{AgNews} \{5.65, 4.14, 4.01, 9.68\};
\textsc{Yelp}: \{60.61, 23.86, 9.15, 33.18\}; 
SNLI: \{12.96, 8.27, 15.04, 6.40\}. 
} 
% \marcos{SNLI uses the decomposable attention classifier}
% \marcos{I see three options here: 1) change the classifier for all datasets to be a rnn with attention as we have before, and we test it with bernoulli and hardkuma attention. 2) we implement a generator-style model for SNLI based on bernoulli and hardkuma selection - this would be more faithful with the Lei et al model. 3) use a bernoulli attention version of the decomposable attention classifier (I'm working on this right now) - this would be more faithful with the Bastings et al model.}
\label{table:results_doc_classification}

\end{table*}

%\input{table_experiment}

%\input{table_backup}

% softmax, sparsemax, entmax: mean (std)
% agnews
% 38.05 (14.31)
% 4.14 (1.95)
% 5.65 (2.41)
% imdb
% 281.42 (208.78)
% 12.94 (3.94)
% 28.23 (10.11)
% snli
% 15.32 (6.69)
% 8.27 (2.62)
% 12.96 (4.93)
% sst
% 19.23 (8.92)
% 2.59 (1.02)
% 4.65 (2.07)
% yelp
% 130.75 (123.80)
% 23.86 (8.71)
% 60.61 (30.91)
% iwslt
% 19.64 (12.69)
% 3.76 (0.77)
% 8.24 (2.15)

\paragraph{Layperson $L$ and explainer $E$.} 
We used a simple linear BoW model as the layperson $L$. 
%For text classification, the layperson only sees the words of the message, without any order information. 
For NLI, 
the layperson sees the full hypothesis, encoding it with a BiLSTM. The BoW from the explainer is passed through a linear projection and summed with the last hidden state of the BiLSTM. % before the output layer.
%We define the messages $m \in \mathcal{M}$ transmitted by the explainer $E$ as bag of words, which is a sparse subset of the original words of $\mathbf{x}$, as described in \S\ref{sec:tasks}.

We evaluated the following explainers: 
 \begin{enumerate}
    \item \textbf{Erasure}, a wrapper  similar to the leave-one-out approaches of  \citet{jain2019attention} and  \citet{serrano2019attention}. We obtain the word with largest attention, zero out its input vector, and repass the whole input with the erased vector to the classifier $C$. % and mask out the erased words in the attention computation. 
    We produce the message by repeating this procedure $k$ times. %and storing the erased words. %The message is a bag with the $k$ recovered words.
    %This strategy is very similar to the leave-one-out experiments from \citet{jain2019attention}, with the exception that we did not remove the feature but replaced it by a null representation. It is also similar to the decision flips experiments from \citet{serrano2019attention}, but instead of re-running the top part of the classifier and look for decision flips, we re-encode the entire input again looking for changes in the attention distribution.
    % \andre{can we also report the sparsemax/entmax attention results, and the average message lengths, without capping to $k$ words? this would be the truly embedded method}
    % \marcos{Yes!}
    \item \textbf{Top-$k$ gradients}, a filter approach that ranks word importance by their  ``input $\times$ gradient'' product,  $|\frac{\partial \hat{y} } { \partial \mathbf{x}_i } \cdot \mathbf{x}_i|$ \citep{ancona2018towards,wiegreffe2019attention}. The top-$k$ words are selected as the message.
    % \marcos{Andre, falar mais sobre o input * gradient? falar que eh bastante usado por outros trabalhos e que consideram como um groundtruth?} \andre{maybe leave it to the related work. there is not much space here.}
    \item \textbf{Top-$k$ and selective attention:} %we use attention probabilities as a measure of importance. %Assuming a softmax function ($\alpha=1$) is used to map attention scores to probabilities, then all words are going to have a nonzero probability mass, implying a tiny but yet existing relevance. Therefore, we resort to cases where $\alpha = \{1.5, 2\}$ to automatically select relevant features through sparsity. 
    We experimented both using attention as a {\it filter}, by selecting the top-$k$ most attended words as the message, and {\it embedded} in the classifier $C$, by using the selective attentions described in \S\ref{sec:attention}  (1.5-entmax and  sparsemax). %, which select $k$ automatically. %When using sparse distributions, we can have cases where less than $k$ words have nonzero probability, for such cases, we simply select all words in the sentence. 
    % Inspired by \citet{serrano2019attention}, we also created an explainer which zero out the maximum attention probability but leave the remaining intact. \andre{this is more an ``adversarial system'', do you think we need to include it? shuffle is another baseline, but why do you call it a ``shuffle softmax''? softmax is not relevant there to produce the explanation, right?} \marcos{I dont think it is like an adversarial system, since is not trained to produce a diff label - and it shows that removing the max prob reduces the score but not that much, which means that attention is focusing on relevant words. The shuffle is because it matters if is sparsemax right? if we shuffle sparsemax we will end up with the same ammount of nonzeros. Thinking about the final result, this will not matter because is the same thing as selecting words at random, so they are pretty much the same, but I wonder if a reviewer is going to complain about this if we dont say the map function used.} 
    %For all of these cases, the message is a bag with the words associated with nonzero probabilities. 
    % We ex periment with $\alpha = \{1.5, 2\}$.  
    % \item \textbf{Encoded attention:} here we also use attention probabilities as a measure of importance, but in this case we selected the top-$k$ words and create a bag of words with them
    \item {\bf The rationalizer models of \citet{lei2016rationalizing} and \citet{bastings2019interpretable}.} These models compose the message by stochastically sampling rationale words, respectively using Bernoulli and HardKuma distributions. For SNLI, since these models use decomposable attention instead of RNNs, we form the message by selecting all premise words that are linked with any hypothesis word via a selected Bernoulli variable. 
    
 \end{enumerate}
%Under the feature selection taxonomy, the erasure explainer can be considered as a wrapper method.
% \andre{why is the random a wrapper? }
% \marcos{it is not, I wrote this very late last night...}
%The top-$k$ attention and gradient explainers can be considered as filter methods. Since sparse attention explainers have a built-in mechanism to select relevant words, they can be regarded as embedded methods.
%We report results for these explainers in the following subsections. 
%As \citet{jain2019attention,serrano2019attention} 
We also report  a {\bf random} baseline, which randomly picks $k$ words as the message. 
%Results for the trainable joint explainer are presented later together with the human evaluation in \S\ref{section:human_evaluation}. 
We show examples of messages for all explainers in~\S\ref{sec:supp_explanations}.

\paragraph{Results.}
Table~\ref{table:results_doc_classification} 
reports results for the communication success rate (CSR, Eq.~\ref{eq:comm_accuracy}) %between the explainer and the layperson (\S\ref{subsec:setup_clf_exp_lay}) 
and for the accuracy of the layperson %on the original task 
(ACC$_L$). For each explainer, we indicate which classifier it is explaining; note that the CSR is only comparable across explainers that use the same classifier. 
The goal of this experiment is to answer the following questions:
(i) How do different explainers (wrappers, filters, embedded) compare to each other?
%\item Does a layperson guided by an explainer perform better than an unguided layperson that sees the entire document?
(ii) Are selective sparse attention methods effective?
(iii) How is the trade-off between message length and CSR? 
\begin{comment}
By comparing the accuracy of the classifiers in Table~\ref{table:results_classifiers} with the ACC$_L$ columns on Table~\ref{table:results_doc_classification}, 
we see a consistent drop from the RNN classifiers to the layperson, regardless of the explainer. This is expected, since the layperson is a much weaker BoW classifier, and it only has access to a limited number of words in the document. Note, however, that for some explainers, this layperson is on par or outperforms a BoW classifier with access to all words. 
\end{comment}

%{\it Attention and erasure outperform gradient, and all outperform random.} 
The first thing to note is that, as expected, the random baseline is much worse than the other explainers, for all text classification datasets.%
\footnote{This is less pronounced in SNLI, as the hypothesis alone already gives strong baselines \citep{gururangan-etal-2018-annotation}.} 
%This means that picking random words as messages does not provide very effective explanations. 
%An exception is {\sc AgNews} where the CSR is still relatively high (92.38\%), though still well below the others. 
Among the non-trivial explainers, 
{\bf the attention and erasure outperform gradient methods}:  
the erasure and top-$k$ attention explainers have similar CSR, with a slight advantage for attention methods. Note that the attention explainers have the important advantage of requiring a single call to the classifier, whereas the erasure methods, being wrappers, require $k$ calls. 
%This is an important advantage if runtime is a concern. 
The worse performance of  top-$k$ gradient  (less severe on {\sc AgNews}) suggests that the words that locally cause bigger output changes are not necessarily the most informative ones.%
\footnote{A potential reason is that attention directly influences $C$'s decisions, being an inside component of the model. Gradients and erasure, however, are extracted after decisions are performed. The reason might be similar to filter methods being generally inferior to embedded methods in static feature selection, since they ignore feature interactions that may jointly play a role in model's decisions.} 

Regarding the different attention models  (softmax, entmax, and sparsemax), we see that  {\bf sparse transformations tend to have slightly better ACC$_L$}, in addition to better ACC$_C$ (see Table~\ref{table:results_classifiers}). % %\andre{regarding CSR, we have to note that $C$ is not the same for all, though}
\remove{
\footnote{For {\sc AgNews}, simply selecting the first $k$ words of the text yields very high CSR: 93.92 and ACC: 92.49. This is consistent with prior findings by \citet{wiegreffe2019attention}.}
} 
The embedded sparse attention methods achieved communication scores on par with the top-$k$ attention methods without a prescribed $k$, while producing, by construction, more faithful explanations. Both our proposed models (sparsemax and 1.5-entmax) seem generally more accurate than the Bernoulli model of \citet{lei2016rationalizing} and 
comparable to the HardKuma model of \citet{bastings2019interpretable}, with a much simpler training procedure, not requiring gradient estimation over stochastic computation graphs. 
%Note, however, that these methods may produce messages longer than $k$ words, hence the comparison with the top-$k$ methods is not completely fair. 
%It is also noticeable that, on the other hand, the top-$k$ attention methods achieved better layperson accuracy. 

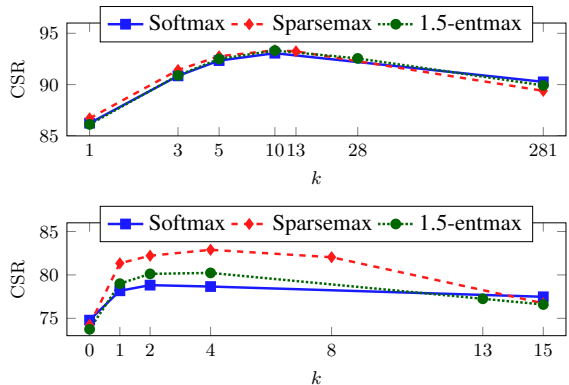
\begin{figure}[tb]
% for now i'll let the figure this way, if I find time i'll search how to use only one legend
% it is not easy to do this and have two subfigures
    \centering
    
    % IMDB
    % CSR plot with subtitle
    % \subfloat[IMDB]{{\input{figs/line_imdb_csr.tex}}}
    % CSR plot
    \subfloat{{\begin{tikzpicture}[scale=0.78]
\begin{axis}[
    cycle list name=my black white,
    % cycle list name=black white,
    % smooth,
    xlabel={$k$},
    xmode=log,
    log ticks with fixed point,
    % x filter/.code=\pgfmathparse{#1 + 6.90775527898214},
    % axis y line=left,
    ylabel={$\mathrm{CSR}$},
    ymin=85, ymax=96,
    enlarge x limits=0.05,
    legend cell align={left},
    xtick={1,3,5,10,13,28,281},
    ytick={85,90,95},
    % legend pos=south east,
    legend style={at={(0.5,1.15)}, anchor=north,legend columns=-1,font=\normalsize},
    % ymajorgrids=true,
    every axis plot/.append style={very thick},
    tick label style={font=\footnotesize},
    % max space between ticks={50},
    % log basis x={20},
    % xtick placement tolerance={25.2},
    % try min ticks log={5},
    % try min ticks={1},
    % minor tick num={1},
    % width=11cm,
    height=3.5cm,
    font=\small,
]

\addplot
coordinates {
    (1,86.24)    (3,90.85)   (5,92.34)   (10,93.06)   
    (281,90.27) % 89.59 
};

\addplot
coordinates {
    (1,86.71)    (3,91.42)   (5,92.76)   (10,93.34)   
    (13,93.24)  
    (281,89.39)   
};

\addplot
coordinates {
    (1,86.12)    (3,90.89)   (5,92.50)   (10,93.32)   
    (28,92.55)   
    (281,89.92)   
}; 

\legend{Softmax, Sparsemax, 1.5-entmax}

\end{axis}
\end{tikzpicture}}}
    % ACC_L plot
    % \subfloat{{\input{figs/line_imdb.tex}}}
    % \quad
    \vspace{-0.2cm}
    % SNLI
    % CSR plot with subtitle
    % \subfloat[SNLI]{{\input{figs/line_snli_csr.tex}}}
    % CSR plot
    \subfloat{{\begin{tikzpicture}[scale=0.78]
\begin{axis}[
    cycle list name=my black white,
    xlabel={$k$},
    % xmode=log,
    % log ticks with fixed point,
    % x filter/.code=\pgfmathparse{#1 + 6.90775527898214},
    % axis y line=left,
    ylabel={$\mathrm{CSR}$},
    ymin=73, ymax=86,
    enlarge x limits=0.05,
    legend cell align={left},
    xtick={0,1,2,4,8,13,15},
    % symbolic x coords={1,3,5,10,13,28,280},
    ytick={75,80,85},
    % legend pos=north east,
    legend style={at={(0.5,1.15)}, anchor=north,legend columns=-1,font=\normalsize},
    % ymajorgrids=true,
    every axis plot/.append style={very thick},
    tick label style={font=\footnotesize},
    % max space between ticks={50},
    % log basis x={2},
    % xtick placement tolerance={25.2},
    % try min ticks log={5},
    % try min ticks={1},
    % minor tick num={1},
    % width=2cm,
    % width=\columnwidth,
    height=3.5cm,
    font=\small,
]

\addplot
coordinates {
    (0,74.79)    (1,78.17)   (2,78.82)   (4,78.66)   (15,77.48)
};

\addplot
coordinates {
    (0,74.24)    (1,81.34)   (2,82.21)   (4,82.89)  (8,82.04)    (15,76.73)
};

\addplot
coordinates {
    (0,73.74)    (1,78.99)   (2,80.12)   (4,80.23)   (13,77.25)  (15,76.59)
}; 

\legend{Softmax, Sparsemax, 1.5-entmax}

\end{axis}
\end{tikzpicture}}}
    % ACC_L plot
    % \subfloat{{\input{figs/line_snli.tex}}}
    % IWSLT
    % CSR plot
    %\subfloat{{\input{figs/line_iwslt_csr.tex}}}
    \caption{Message sparsity analysis for IMDB (top) and SNLI (bottom). For SNLI, $k=0$ corresponds to a case where the layperson only sees the hypothesis.
    The rightmost entry represents an explainer that simply passes forward all words to the layperson. %The average $k$ for sparsemax and 1.5-entmax are, respectively: 13 and 28 for IMDB; 8 and 13 for SNLI.
    % \andre{maybe show ACC plots as well? eg ACC of L as function of $k$ and flat line with ACC of C for each model} 
    % \marcos{In the same plot or in a new figure? ps: we discussed last week about whether these plots are misleading or not because each line represents a different classifier.} 
    } \label{fig:k_analysis}
    %\andre{consider reporting only SNLI.}
    %\marcos{Due to lack of space?}
    % Say something about the performance of $k=0$ in SNLI}
    
\end{figure}

Finally, Figure~\ref{fig:k_analysis} shows the trade-off between the length of the message and the communication success rate for different values of $k$ both for IMDB and SNLI (see Figure~\ref{fig:k_analysis_nmt} in \S\ref{sec:nmt_setup} for the IWSLT experiments, with similar findings). 
Interestingly, we observe that 
{\bf CSR does not increase monotonically with $k$.} 
% To better understand the trade-off between the length of the message and the communication success rate, we plot in Figure~\ref{fig:k_analysis} how different systems compare for different values of $k$ both for IMDB (text classification) and SNLI (natural language inference). 
As $k$ increases, CSR starts by increasing but then it starts dropping when $k$ becomes too large. 
This matches our intuition: in the two extreme cases where $k=0$ and where $k$ is the document length (corresponding to a full bag-of-words classifier) the message has no information about how the classifier  $C$ behaves. 
%\marcos{Remove: This behaviour is more clear for SNLI, where the average length of the premise is of 15 words only.} 
By setting $k=0$, meaning that the layperson $L$ only looks at the hypothesis, the $\mathrm{CSR}$ is reasonably high (\til74\%), but as soon as we include a single word in the message this baseline is surpassed by 4 points or more. 

\section{Human Evaluation} \label{sec:human_evaluation}

To fully assess the quality of the explanations in a more realistic forward simulation setting, we performed human evaluations, where the layperson $L$ is a human instead of a machine.

\begin{table*}[t]
\small
\begin{center}
% \resizebox{\textwidth}{!}{%
\begin{tabular}{ll ccccc c ccccc}
\toprule
\multirow{2}{*}{\sc Clf.} &
\multirow{2}{*}{\sc Explainer} & \multicolumn{5}{c}{IMDB} & & \multicolumn{5}{c}{SNLI} \\
\cmidrule{3-7}
\cmidrule{9-13}
        & & $k$ & CSR$_H$ & CSR$_L$ & ACC$_H$ & ACC$_L$         & & $k$ & CSR$_H$ & CSR$_L$ & ACC$_H$ & ACC$_L$ \\    
\midrule 

$C$ & Erasure 	                        & 5.0 & 89.25 & 94.00 & 86.25 & 90.00 		& & 4.0 & 72.50 & 73.50 & 83.50 & 70.00   \\
$C$ & Top-$k$ gradient 					& 5.0 & 73.50 & 84.50 & 73.00 & 80.50 	    & &	4.0 & 65.75 & 72.50 & 76.75 & 68.00    \\
$C$ & Top-$k$ softmax 					& 5.0 & 89.25 & 93.00 & 88.25 & 88.00 	    & &	4.0 & 72.00 & 76.50 & 82.75 & 71.50   \\
$C_{\text{ent}}$ & Top-$k$ 1.5-entmax 	& 5.0 & 89.25 & 92.50 & 85.75 & 86.50  	    & &	4.0 & 70.00 & 81.50 & 80.50 & 76.50   \\
$C_{\text{sp}}$ & Top-$k$ sparsemax 	& 5.0 & 89.00 & 89.50 & 87.50 & 88.00  	    & &	4.0 & 68.25 & 88.00 & 80.25 & 77.00   \\
\midrule
$C_{\text{ent}}$ & Selec. 1.5-entmax 	& 27.2 & 86.50 & 92.50 & 84.00 & 89.50 		& & 12.9 & 75.25 & 77.00 & 87.00 & 77.00   \\
$C_{\text{sp}}$ & Selec. sparsemax 		& 12.8 & 87.75 & 92.50 & 86.75 & 89.00 		& & 8.0  & 72.25 & 82.00 & 85.00 & 79.00   \\
% $C_{\text{bern}}$ & Bernoulli 			& 57.9 & 70.75 & 92.00 & 70.75 & 86.00 		& & 15.2 & 74.50 & 76.00 & 86.75 & 69.50  	\\
% $C_{\text{hk}}$ & HardKuma 				& 67.2 & 67.00 & 92.00 & 68.00 & 86.50 		& &	6.4  & 79.25 & 71.50 & \bf 87.50 & 68.50 	\\
$C_{\text{bern}}$ & Bernoulli 			& 39.4 & 79.00 & 93.50 & 75.00 & 87.00 		& & 15.2 & 74.50 & 76.00 & 86.75 & 69.50  	\\
$C_{\text{hk}}$ & HardKuma 				& 24.3 & 83.75 & 93.50 & 80.75 & 89.00 		& &	6.4  & 79.25 & 71.50 & \bf 87.50 & 68.50 	\\
\midrule
$C$ & Joint $E$ and $L$ 				& 2.7  & \bf 96.75 & \bf 98.50 & \bf 89.25 & \bf 91.50 	    & &	2.8 & 58.00 & \bf 93.50 & 70.00 & 78.50 	\\
- & Human highlights					& -  & -     & -     & -     & -                            & &	2.8 & \bf 83.25  & 83.50 & 83.25 & \bf 83.50 \\
% - & Human highlights					& -  & -     & -     & -     & -      	& &	3 & 69.00 & 76.00 & \bf 86.50 & 77.00 \\
% H CSR and L CSR nao faz muito sentido!
\bottomrule
\end{tabular}
% }
\end{center}
\caption{Results of the human evaluation. Reported are average message length $k$, human layperson CSR$_H$/ACC$_H$, and machine layperson  CSR$_L$/ACC$_L$. 
% \andre{need to report message size of non-top-$k$ methods, and for human highlights we should have $CSR=ACC$}
Only explainers of the same classifier can be compared in terms of CSR.
} 
\label{table:results_human}
\end{table*}

\paragraph{Joint training of $E$ and $L$.} 
% \andre{maybe put this here? an alternative is to put it together with the human evaluation section} 
%The CSR can be viewed as a  metric for explanation quality in terms of the communication between $E$ and $L$. But 
So far we compared several explainers, but what happens if we train $E$ and $L$ jointly to optimize CSR directly, as described in \S\ref{subsec:joint_e_and_l}? %Here, we also evaluate a trainable post-hoc explainer $E$ that is trained to produce good explanations by selecting a subset of relevant words in a end-to-end fashion as described in . 
%This means that the layperson's loss gradient is backpropagated through the network to adjust the explainer's parameters and therefore a high CSR is expected. 
We experiment with the IMDB and SNLI datasets, %following the setup described in \S\ref{subsec:joint_e_and_l}, 
%with $\lambda=1$ and $\beta=20\%$ for both, chosen in the validation set as described in \S\ref{sec:analysis_beta}. 
% \andre{update this with the final hyperparameters}
comparing with using humans for either the layperson, the explainer, or both. 

%Since top-$k$ is not a differentiable operation, we use sparsemax attention during training, to compose the explainer's message. The message is a normalized bag of probabilities during training, but it is still a top-$k$ bag-of-words during test time. 
% We let the explainer access the predictions of the classifier ($\hat{y}$) during training by following a linear increasing schedule, so that at the end of training the explainer has a probability of 20\% of accessing $\hat{y}$. %We set $\lambda=2$ for IMDB and $\lambda=20$ for SNLI.
%\marcos{Nao sei se faz sentido isso sobre otimizar a comunicacao, mas me parece um ponto valido}
%The experiments with this explainer aims that answering the question: can we extract meaningful explanations by ``optimizing'' the communication?
%\marcos{Se colocar essa questao temos que responde-la depois}

\paragraph{Human layperson.} 
%An important aspect of our framework is how it relates to human judgement. 
%\andre{we may need to cite \url{https://dongnguyen.nl/publications/nguyen-naacl2018.pdf} somewhere} 
% We carried all explainers used in the previous experiments 
% \andre{maybe all?}
% \marcos{it's missing only the erasure explainer from the table - should I include it in the human annotation?} \andre{if you can, it would be great}
% and 
We randomly selected 200 documents for IMDB and SNLI to be annotated by humans. %\toremove{The documents were displayed in a different order for each explainer.} \toremove{Since in our experimentfs we define the message as being a bag-of-words, which does not encode order information,} 
The extracted explanations (i.e. the selected words) were shuffled and displayed as a cloud of words to two annotators, who were asked to predict the label of each document when seeing only these explanations. 
% \andre{can we report inter-annotator agreement?} 
% \marcos{sure! does the average agreement from all explainers sounds good?}
For SNLI, we show the entire hypothesis as raw text and the premise as a cloud of words.
The agreement between annotators and other annotation details can be found in \S\ref{sec:supp_interface}.

\paragraph{Human explainer.} 
% \andre{need to trim substantially}
We also consider explanations generated by humans rather than machines. To this end, we used the e-SNLI corpus \citep{esnli2018}, which extends the SNLI with human rationales. 
% We dropped 25 examples in the training set of e-SNLI due to missing explanations. 
Since the e-SNLI corpus does not provide highlights over the premise for neutral pairs, 
% , and therefore our previous explainers for SNLI can not be directly compared. 
we removed them from the test set.%
%and obtained explanations using our trained models 
\footnote{Note that the human rationales from eSNLI are not explanations about $C$, since the humans are explaining the gold labels. Therefore, we have CSR$=$ACC always.}

\smallskip

We summarize our results in Table~\ref{table:results_human}. 
We observe that, also with human laypeople, {top-$k$ attention  achieves better results than top-$k$ gradient},  in terms of CSR and ACC, 
% With the exception of top-$k$ sparsemax, in IMDB we see that the machine-learned $L$ performs better than humans. For SNLI the opposite happens, with the exception for the layperson trained with the joint explainer. 
and that 
{the ACC of erasure, attention models, and human explainers are close}, reinforcing again the good results for these explainers. 
%Although we report both CSR$_H$ and CSR$_L$ for the human highlights explainer, we point out that humans are explaining the gold labels and not the classifier's $C$ decisions, which may explain the relatively low scores. %Nonetheless, is interesting to see that the CSR for this explainer is higher than the top-$k$ gradient explainer. 
%\andre{We can also compare the word overlap in the explanations among the human and automatic explainers. If the overlap is big, the performance of the laypeople will be similar.}
%\marcos{This is similar with the precision analysis from (Lei et al), right? I like it, but dont know if it fits here though.}
%Besides human highlights, eSNLI also has natural language explanations for a pair label. We found that by using this natural language explanations as message, the performance of the layperson grows to CSR of 80.54 and ACC of 85.29, which is consistent with the very strong results found originally in \citet{esnli2018}. 
% \andre{Claim: sparse attention better than dense one - can only claim this for ACC!}
Among the different attention explainers, we  see that selective attention explainers (\S\ref{sec:attention}) got very high ACC$_H$, outperforming top-$k$ explainers for SNLI. 
We also see that {the joint explainer (\S\ref{subsec:joint_e_and_l}) outperformed all the other explainers in ACC$_L$ and  CSR$_L$  
% \andre{note: cannot compare with sparse attentions or human highlights in CSR, since the classifier is different} 
%The large values of CSR$_L$ are expected, since this explainer was trained to optimize this metric. 
%As expected, the joint explainer achieves very high $L$-CSR, also outperforming all explainers in terms of $L$-ACC. % \andre{we should emphasize this more!}. 
%\marcos{Yes but carefully. I'd say that the explanations extracted from IMDB are ok, but it is not the case for SNLI. We should first have a metric that dictates the quality of the explanations. Maybe, as you suggested, the second loss term?}
%\andre{Claim: same for H-ACC on IMDB, but not for SNLI}
and achieved  very high human performance on IMDB, largely surpassing other systems in CSR$_H$ and ACC$_H$. 
This shows the potential of our communication-based framework to develop new post-hoc explainers with good forward simulation properties. 
However, for SNLI, the joint explainer had much lower CSR$_H$ and ACC$_H$, suggesting that for this task more sophisticated explainers are required. %

\section{Related Work}\label{sec:related_work}

%The area of explainable AI is growing very fast, with

%Recent work used the information bottleneck \citep{} minimum description principle \citep{} 
%\andre{todo: add relation to MDL principle (Voita's paper) and IB principle (Eisner's paper, see \url{https://medium.com/jasonwu0731/information-bottleneck-for-nlp-parsing-summarization-961418fbb697})}

There is a large body of work on analysis and interpretation of neural networks. 
%One area of research, not addressed in our paper, concerns understanding what different neural architectures can and cannot do, and  distilling them to models that are globally more amenable to interpretation, such as finite-state automata \citep{weiss2018practical,weiss2018extracting,schwartz2018bridging,peng2018rational}. 
Our work focuses on {\it prediction explainability}, different from transparency or model interpretability \citep{doshi2017towards,lipton2016mythos,gilpin2018explaining}.  %,riedl2019human}. 

\citet{rudin2019stop} defines explainability as a plausible reconstruction of the decision-making process, and \citet{riedl2019human} argues that it mimics what humans do when rationalizing past actions. This inspired our post-hoc explainers in \S\ref{subsec:joint_e_and_l} and their use of the faithfulness loss term. 

%The area of explainable AI is growing very fast. Several terminologies have been proposed in the last few years in order to have concise definitions of explainable AI properties \citep{lipton2016mythos,doshi2017towards}. 
%One option to obtain transparency is by developing models that are natively interpretable, such as ... \citep{rudin2019stop}, rather than attempting to explain non-transparent models. Whether this can be done without sacrificing model accuracy is currently unclear.

Recent works questioned the interpretative ability of attention mechanisms %and whether they can be regarded as a form of explanation or not
\citep{jain2019attention,serrano2019attention}. 
\citet{wiegreffe2019attention} distinguished between faithful and plausible explanations and introduced several diagnostic tools. 
%One of their tests ({\it diagnosing attention distributions by guiding simpler models}, in their  \S3.4) has similarities with the relation between the explainer and the layperson proposed in our work. 
%\citet{jain2019attention} claims attention is not explanation by comparing attention mappings to gradient probing information. \citet{serrano2019attention} proposes a ablation strategy that eliminates one feature at the time looking for decision shifts, concluding that attention is not a fail-safe indicator. \citet{wiegreffe2019attention} questions the conclusions of \citet{jain2019attention} and proposes various explainability tests. 
%These works, however, resort to analyzing attention-based methods, and 
\citet{mullenbach-etal-2018-explainable}  use human evaluation to show that attention mechanisms produce plausible explanations, consistent with our findings in \S\ref{sec:human_evaluation}. 
None of these works, however, considered the sparse selective attention mechanisms  proposed in \S\ref{sec:attention}. 
%Our goal is not to distinguish between faithful and plausible explanations. In contrast, we focus on better characterizing when explanations should be considered plausible, by providing a framework that gives an objective answer to this question. 
%As discussed in \citet[\S5]{wiegreffe2019attention}, plausible explanations are very important even if not faithful: \citet{rudin2019stop} defines explainability as a plausible reconstruction of the decision-making process, and \citet{riedl2019human} argues that they mimic what humans do when rationalizing past actions. 
Hard stochastic attention has been considered by  \citet{xu2015show,lei2016rationalizing,alvarez2017causal,bastings2019interpretable}, but a  comparison with sparse attention and explanation strategies was still missing.

%\andre{we may need to justify somewhere why we don't follow their proposal of examining the prediction power of attention distributions in a ‘clean’ setting, where the trained parts of the model have no access to neighboring tokens of the instance.} 

Besides attention-based methods, 
many other explainers have been proposed using gradients \citep{bach2015pixel,Montavon2018,ding-etal-2019-saliency}, %\citep{bach2015pixel}, DeepLIFT \citep{shrikumar2017learning} and Integrated Gradients \citep{sundararajan2016gradients}, 
leave-one-out strategies \citep{feng2018pathologies,serrano2019attention}, 
or local perturbations  \citep{ribeiro2016should,Koh2017}, but a link with filters and wrappers in the  feature selection literature has never been made. 
We believe the connections revealed in  \S\ref{sec:feature_selection} may be useful to develop new explainers in the future.

Our trained explainers from \S\ref{subsec:joint_e_and_l} draw inspiration from emergent communication \citep{lazaridou2016multi,foerster2016learning,havrylov2017emergence}. Some of our proposed ideas (e.g., using sparsemax for end-to-end differentiability) may also be relevant to that task. 
Our work is also related to sparse auto-encoders, which seek sparse overcomplete vector representations to improve model interpretability \citep{faruqui2015sparse,trifonov2018learning,subramanian2018spine}. 
In contrast to these works, we consider the non-zero attention probabilities as a form of explanation.

Some recent work \citep{yu2019rethinking,deyoung2019eraser} advocates {\it comprehensive} rationales. 
While comprehensiveness could be useful in our framework to prevent trivial communication protocols between the explainer and layperson, 
we argue that it is not always a desirable property, since it leads to longer explanations and an increase of human cognitive load. In fact, our analysis of CSR as a function of message length (Figure~\ref{fig:k_analysis}) suggests that shorter explanations might be preferable. This is aligned to the ``explanation selection'' principle articulated by \citet[\S 4]{miller2019explanation}: {\it ``Similar to causal connection, people do not typically provide all causes for an event
as an explanation. Instead, they select what they believe are the most relevant causes.''}
Our sparse, selective attention mechanisms proposed in \S\ref{sec:attention} are inspired by this principle.  

\section{Conclusions}

We proposed a unified framework that regards explainability as a communication problem between an explainer and a layperson about a classifier's decision. 
We proposed new embedded methods based on selective attention, and post-hoc  explainers trained to optimize communication success. 
In our experiments, we observed that attention mechanisms and erasure tend to outperform gradient methods on communication success rate, 
using both machines and humans as the layperson, and that selective attention is effective, while simpler to train than stochastic rationalizers. 

%\marcos{Since our framework is agnostic about the model used for the original classifier, it would be interesting to use higher complexity models, such as Transformers. 
%On the other hand, Transformers would make it harder to assess the validity of our framework since it has multiple attention heads and layers. 
%We also plan to evaluate our framework on other tasks that heavily rely on attention mechanism, such as language modelling and speech recognition.}

\section*{Acknowledgements}
This work was supported by the European Research Council (ERC StG DeepSPIN 758969), by the P2020 program MAIA (contract 045909), and by the Fundação para a Ciência e Tecnologia through contract UID/50008/2019. We are grateful to Thales Bertaglia, Erick Fonseca, Pedro Martins, Vlad Niculae, Ben Peters, Gonçalo Correia and Tsvetomila Mihaylova for insightful group discussion and for the participation in human evaluation experiments. We also thank the anonymous reviewers for their helpful discussion and feedback.

% \bibliography{anthology,emnlp2020}
\bibliography{emnlp2020}
\bibliographystyle{acl_natbib}

\appendix

\clearpage
\onecolumn
\appendix

% \section{Supplemental Material}
% \label{sec:supplemental}

\section{Sparse attention}\label{sec:supp_attention}

% \andre{need to modify below}
% An example is the {\bf sparsemax transformation} \citep{martins2016softmax}, 
A natural way to get a sparse attention distribution is by using the {\bf sparsemax transformation} \citep{martins2016softmax},
which computes an Euclidean projection of the score vector onto the probability simplex $\triangle^n := \{\mathbf{p} \in \mathbb{R}^n \mid \mathbf{p}\ge \mathbf{0}, \,\, \mathbf{1}^\top\mathbf{p} = 1\}$, or, more generally, the {\bf $\alpha$-entmax transformation} \citep{peters2019sparse}: 
    \begin{equation}\label{eq:entmax}
        \alpha\text{-entmax}(\mathbf{s}) := \argmax_{\mathbf{p} \in \triangle^{n}} \mathbf{p}^\top \mathbf{s} + H_\alpha(\mathbf{p}),
    \end{equation}
where $H_\alpha$ is a generalization of the Shannon and Gini entropies proposed by \citet{Tsallis1988},  parametrized by a scalar $\alpha\ge 1$: 
\begin{equation}\label{eq:tsallis}
    H_\alpha (\mathbf{p}) := \begin{cases}
                                                \frac{1}{\alpha(\alpha-1)}\sum_j(p_j-p_j^\alpha), &  \alpha \neq 1\\
                                                -\sum_j p_j \log p_j, & \alpha=1.
                                            \end{cases}
\end{equation}
Setting $\alpha=1$ recovers the softmax function, while for any value of $\alpha>1$ this transformation can return a sparse probability vector. %(as the value of $\alpha$ increases, the induced probability distribution becomes more sparse). 
Letting $\alpha=2$, we recover sparsemax. A popular choice is $\alpha=1.5$, which has been successfully used  in machine translation and morphological inflection applications \citep{peters2019sparse}.

\section{Data statistics and preparation} \label{sec:data_statistics_preparations}

% We picked the same datasets as \citet{jain2019attention} and \citet{wiegreffe2019attention}, excluding the smallest ones. Like them, for AgNews, we considered the binary case of World vs Business articles. 
We used four datasets for text classification:  
SST,\footnote{\url{https://nlp.stanford.edu/sentiment/}} 
IMDB,\footnote{\url{https://ai.stanford.edu/~amaas/data/sentiment/}}
AgNews,\footnote{\url{https://www.di.unipi.it/~gulli/AG_corpus_of_news_articles.html}} 
and Yelp.\footnote{\url{https://www.yelp.com/dataset/}}
One dataset for NLI:
SNLI,\footnote{\url{https://nlp.stanford.edu/projects/snli/}} along with its extended version (eSNLI\footnote{\url{https://github.com/OanaMariaCamburu/e-SNLI}}) which includes human-annotated explanations of the entailment relations \citep{esnli2018}. And the EN$\to$DE IWSLT 2017 dataset for machine translation \citep{cettolo2017overview}.\footnote{\url{https://wit3.fbk.eu/mt.php?release=2017-01-trnted}} Table~\ref{table:datasets} shows statistics for each dataset.

\begin{table}[!htb]
    \small
    \begin{center}
    \begin{tabular}{lcccc}
        \toprule
        \sc Name & \sc \# Train & \sc \# Test & \sc Avg. tokens & \sc \# Classes \\
        \midrule
        SST         & 6920          & 1821          & 19          & 2 \\
        IMDB        & 25K           & 25K           & 280         & 2 \\
        AgNews     & 115K          & 20K           & 38          & 2 \\
        Yelp        & 5.6M          & 1M            & 130         & 5 \\
        SNLI        & 549K          & 9824          & 14 / 8      & 3 \\
        % MNLI        & 392K          & 20K         & 22          & 3 \\
        %eSNLI       & 549K          & 9824          & 14 / 8      & 3 \\
        IWSLT       & 206K          & 2271          & 20 / 18     & 134,086  \\
        % source vocab size: 58728       
        % target vocab size: 134086
        \bottomrule
    \end{tabular}
    \end{center}
    \caption{Dataset statistics. The average number of tokens for SNLI is related to the premise and hypothesis, and for IWSLT to the source and target sentences.} 
    \label{table:datasets}
\end{table}

For AgNews, we considered the binary case of World vs Business articles. Although the selected datasets are the same as in previous works \citep{jain2019attention,wiegreffe2019attention}, the training and test set might differ. For SST, IMDB, IWSLT and SNLI  we used the standard splits, but for AgNews and Yelp we randomly split the dataset, leaving 85\% for training and 15\% for test. 
Moreover, for IMDB, AgNews and Yelp we randomly selected 10\%, 15\% and 15\% of examples from the training set to be used as validation data, respectively.

\section{Computing infrastructure}

Our infrastructure consists of 4 machines with the specifications shown in Table~\ref{table:computing_infrastructure}. The machines were used interchangeably, and all experiments were executed in a single GPU. Despite having machines with different specifications, we did not observe large differences in the execution time of our models across different machines. 
% Furthermore, all of our models fit in a single GPU. 

\begin{table}[!htb]
    \small
    \begin{center}
    \begin{tabular}{l ll}
        \toprule
        \sc \# & \sc GPU & \sc CPU  \\
        \midrule
        % hermes
        1.   & 4 $\times$ Titan Xp - 12GB           & 16 $\times$ AMD Ryzen 1950X @ 3.40GHz - 128GB \\
        % athena
        2.   & 4 $\times$ GTX 1080 Ti - 12GB        & 8 $\times$ Intel i7-9800X @ 3.80GHz - 128GB \\
        % zeus
        3.   & 3 $\times$ RTX 2080 Ti - 12GB        & 12 $\times$ AMD Ryzen 2920X @ 3.50GHz - 128GB \\
        % hera
        4.   & 3 $\times$ RTX 2080 Ti - 12GB        & 12 $\times$ AMD Ryzen 2920X @ 3.50GHz - 128GB \\
        \bottomrule
    \end{tabular}
    \end{center}
    \caption{Computing infrastructure.} 
    \label{table:computing_infrastructure}
\end{table}

\section{Classifiers experimental setup (Table~\ref{table:results_classifiers})} \label{sec:training_details}

We chose our classifiers so that they are close to the models used by related works \citep{jain2019attention,wiegreffe2019attention,bastings2019interpretable}. For all models, we calculated their accuracy on the dev set after each epoch. At the end of training we selected the model with the best validation accuracy. We experimented with two classes of classifiers: a simple RNN with attention as in \citet{jain2019attention,wiegreffe2019attention}; and the rationalizer models of \citet{lei2016rationalizing} and \citet{bastings2019interpretable} which sample binary masks from Bernoulli and HardKuma distributions, respectively. 

\subsection{RNNs with attention} \label{subsec:rnns_with_attn}
For the text classification experiments, each input word $x_i$ is mapped to 300D-pretrained GloVe embeddings \citep{pennington2014glove} from the 840B release,\footnote{\url{http://nlp.stanford.edu/data/glove.840B.300d.zip}} kept frozen, followed by a bidirectional LSTM layer (BiLSTM) resulting in vectors $\mathbf{h}_1,\ldots,\mathbf{h}_n$. 
We score each of these vectors using the additive formulation of \citet{Bahdanau2015}, applying an attention transformation to convert the resulting scores $\mathbf{s} \in \mathbb{R}^n$ to a probability distribution $\pi \in \triangle^n$. 
We use this to compute a  contextual vector $\mathbf{c} = \sum_{i=1}^n \pi_i \mathbf{h_i}$, which  is  fed into the output softmax layer that predicts $\hat{y}$. 
For NLI, the input $x$ is a pair of sentences (a premise and an hypothesis), and the classifier $C$ is similar to the the above, but with two independent BiLSTM layers, one for each sentence. In the attention layer, we use the last hidden state of the hypothesis as the query and the premise vectors as keys. 

We used the AdamW \cite{loshchilov2018decoupled} optimizer for all experiments. We tuned two hyperparameters: learning rate within $\{0.003, \mathbf{0.001}, 0.0001\}$, and $l_2$ regularization within $\{0.01, 0.001, \mathbf{0.0001}, 0\}$. We picked the best configuration by doing a grid search and by taking into consideration the accuracy on the validation set (selected values in bold).  Table~\ref{tab:table_all_hyperparams_rnns} shows all hyperparameters set for training.

\begin{table}[!htb]
    \centering
    \small
    \begin{tabular}{llllll}
    \toprule
    \sc Hyperparam.             & \sc SST & \sc IMDB & \sc AgNews & \sc Yelp &  \sc SNLI \\
    \midrule
    Word embeddings size        & 300   & 300   & 300   & 300   & 300      \\
    BiLSTM hidden size          & 128   & 128   & 128   & 128   & 128     \\
    Merge BiLSTM states         & concat    & concat    & concat    & concat    & concat \\
    
    Batch size                  & 8     & 16    & 16    & 128   & 32      \\
    Number of epochs            & 10    & 10    & 5     & 5    & 10    \\
    Early stopping patience     & 5     & 5     & 3     & 3     & 5     \\
    Learning rate               & 0.001     & 0.001     & 0.001     & 0.001     & 0.001     \\
    $\ell_2$ regularization     & 0.0001    & 0.0001    & 0.0001    & 0.0001    & 0.0001     \\
    \bottomrule
    \end{tabular}
    \caption{RNNs training hyperparameters for text classification and NLI datasets.}
    \label{tab:table_all_hyperparams_rnns}
\end{table}

\subsection{Bernoulli and HardKuma} \label{subsec:bernolli_and_hardkuma}

We used the implementation of \citet{bastings2019interpretable},\footnote{\url{https://github.com/bastings/interpretable_predictions}} which includes a reimplementation of the generator-encoder model from \citep{lei2016rationalizing}.
The model used for text classification is a RNN-based generator followed by a RNN-based encoder, whereas for NLI is a decomposable attention classifier from \citep{parikh-etal-2016-decomposable}, for which only the HardKuma implementation was available. 
In order to faithfully compare the frameworks, we adapted the HardKuma code and implemented a Bernoulli version of the same classifier, taking into consideration the sparsity and fused-lasso loss penalties, and the deterministic strategy used during test time. For simplicity, we used the independent variant of the generator of \citet{lei2016rationalizing}.
Table~\ref{tab:table_all_hyperparams_rationalizing} lists only the hyperparameters that we set during training. We refer to the original work of \citet{bastings2019interpretable} to see all other hyperparamers, for which we kept the default values.

\begin{table}[!htb]
    \centering
    \small
    \begin{tabular}{llllll}
    \toprule
    \sc Hyperparam.                 & \sc SST & \sc IMDB & \sc AgNews & \sc Yelp &  \sc SNLI \\
    \midrule
    % Latent selection (HardKuma)     & 0.3       & 0.3       & 0.3       & 0.3       & 0.1      \\
    % Sparsity penalty (Bernoulli)    & 0.01      & 0.0003    & 0.01      & 0.01      & 0.0003      \\
    Latent selection (HardKuma)     & 0.3       & 0.1       & 0.3       & 0.3       & 0.1      \\
    Sparsity penalty (Bernoulli)    & 0.01      & 0.001     & 0.01      & 0.01      & 0.0003      \\
    
    Lasso penalty    & 0        & 0     & 0      & 0      & 0      \\
    
    Batch size                      & 25        & 25        & 25        & 256       & 64     \\
    Number of epochs                & 25        & 25        & 25        & 10        & 100    \\
    Early stopping patience         & 5         & 5         & 5         & 5         & 100    \\
    Learning rate                   & 0.0002    & 0.0002    & 0.0002    & 0.001     & 0.0002     \\
    $\ell_2$ regularization         & $10^{-5}$ & $10^{-5}$ & $10^{-5}$ & $10^{-5}$ & $10^{-6}$     \\
    \bottomrule
    \end{tabular}
    \caption{Rationalizer models training hyperparameters for text classification and NLI datasets.}
    \label{tab:table_all_hyperparams_rationalizing}
\end{table}

\subsection{Validation set results and model statistics}

Table~\ref{table:classifiers_statistics_params} shows the accuracy of each classifier on the validation set, their number of trainable parameters and the average training time per epoch.

\begin{table*}[!htb]
\small
\begin{center}
% \resizebox{\textwidth}{!}{%
\begin{tabular}{l ccc c ccc c ccc}
\toprule
& \multicolumn{3}{c}{\sc SST} & & \multicolumn{3}{c}{\sc IMDB} & & \multicolumn{3}{c}{\sc AgNews} \\

\cmidrule{2-4} \cmidrule{6-8} \cmidrule{10-12}

\sc Clf.\,\, & 
\sc \# P & \sc $t$ & \sc ACC  & & 
\sc \# P & \sc $t$ & \sc ACC  & & 
\sc \# P & \sc $t$ & \sc ACC  \\

\midrule 

$C$                 & 474K & 10s & 85.32    & & 474K & 2m & 95.64     & & 474K & 2m & 98.09  \\
$C_{\text{ent}}$    & 474K & 10s & 84.29    & & 474K & 2m & 95.84     & & 474K & 2m & 98.54  \\
$C_{\text{sp}}$     & 474K & 10s & 84.17    & & 474K & 2m & 95.44     & & 474K & 2m & 98.51  \\
$C_{\text{bern}}$   & 1.1M & 15s & 80.16    & & 1.1M & 2m & 87.40     & & 1.1M & 2m & 96.26  \\
$C_{\text{hk}}$     & 1.1M & 15s & 84.40    & & 1.1M & 2m & 91.84     & & 1.1M & 2m & 96.74  \\

\bottomrule
\end{tabular}
% }
\end{center}
\caption{Classifier results on the validation set and model statistics. \# P is the number of trainable parameters, and is $t$ the average training time per epoch.} 

\label{table:classifiers_statistics_params}

\end{table*}

\begin{table*}[!htb]
\small
\begin{center}
% \resizebox{\textwidth}{!}{%
\begin{tabular}{l ccc c ccc}
\toprule
& \multicolumn{3}{c}{\sc Yelp} & & \multicolumn{3}{c}{\sc SNLI} \\

\cmidrule{2-4} \cmidrule{6-8}

\sc Clf.\,\, & 
\sc \# P & \sc $t$ & \sc ACC  & & 
\sc \# P & \sc $t$ & \sc ACC  \\

\midrule 

$C$                 & 474K & 3h & 77.03     & & 998K & 4m & 78.74 \\
$C_{\text{ent}}$    & 474K & 3h & 76.72     & & 998K & 4m & 79.38 \\
$C_{\text{sp}}$     & 474K & 3h & 76.84     & & 998K & 4m & 79.69 \\
$C_{\text{bern}}$   & 1.1M & 5h & 69.99     & & 382K & 2m & 79.79 \\
$C_{\text{hk}}$     & 1.1M & 5h & 74.29     & & 462K & 2m & 86.04 \\

\bottomrule
\end{tabular}
% }
\end{center}
\caption{Continuation of Table~\ref{table:classifiers_statistics_params}.} 

\label{table:classifiers_statistics_params_2}

\end{table*}

\section{Communication experimental setup (Table~\ref{table:results_doc_classification})} \label{sec:communication_details}

Training the communication under our framework consists on training a layperson $L$ on top of explanations (message) produced by $E$ about $C$'s decision. With the exception of the explainer $E$ trained jointly with $L$, none of the other explainers have trainable parameters. Therefore, in these cases, the communication between $E$ and $L$ consists only on training $L$. For all models, we calculated its CSR on the dev set after each epoch. At the end of training we selected the model with the best validation CSR. Table~\ref{tab:table_all_hyperparams_comm} shows the communication hyperparameters. Note that for SNLI we still need to train a BiLSTM to encode the hypothesis.

\begin{table}[!htb]
    \centering
    \small
    \begin{tabular}{llllll}
    \toprule
    \sc Hyperparam.             & \sc SST & \sc IMDB & \sc AgNews & \sc Yelp &  \sc SNLI \\
    \midrule
    Word embeddings size        & -   & -   & -   & -   & 300      \\
    BiLSTM hidden size          & -   & -   & -   & -   & 128     \\
    Merge BiLSTM states         & -   & -   & -   & -   & concat \\
    
    Batch size                  & 16    & 16    & 16    & 112   & 64      \\
    Number of epochs            & 10    & 10    & 10    & 5     & 10    \\
    Early stopping patience     & 3     & 3     & 3     & 3     & 3     \\
    Learning rate               & 0.001 & 0.001 & 0.001 & 0.003 & 0.001     \\
    $\ell_2$ regularization     & $10^{-5}$ & $10^{-5}$ & $10^{-5}$ & $10^{-5}$ & $10^{-5}$     \\
    \bottomrule
    \end{tabular}
    \caption{Communication hyperparameters for text classification and NLI datasets.}
    \label{tab:table_all_hyperparams_comm}
\end{table}

\subsection{Validation set results and model statistics}

Table~\ref{table:communication_statistics_params} shows the CSR and ACC$_L$ for each explainer on the validation set, the number of trainable parameters of $L$ and the average training time per epoch.

\begin{table*}[!htb]
\small
\begin{center}
% \resizebox{\textwidth}{!}{%
\begin{tabular}{l cccc c cccc c cccc}
\toprule
& \multicolumn{4}{c}{\sc SST} & & \multicolumn{4}{c}{\sc IMDB} & & \multicolumn{4}{c}{\sc AgNews}\\

\cmidrule{2-5} \cmidrule{7-10} \cmidrule{12-15}

\sc Explainer\,\, & 
\sc \# P & \sc $t$ & CSR & ACC$_L$  & & 
\sc \# P & \sc $t$ & CSR & ACC$_L$  & & 
\sc \# P & \sc $t$ & CSR & ACC$_L$  \\

\midrule 

Random                  & 38K & 10s & 63.76 & 62.84   & & 247K & 1m & 61.36 & 61.24    & & 120K & 2m & 85.26 & 84.58 \\
Erasure                 & 38K & 10s & 81.88 & 79.82   & & 247K & 2m & 94.00 & 91.40    & & 120K & 3m & 98.41 & 96.98 \\
Top-$k$ gradient        & 38K & 10s & 76.72 & 75.57   & & 247K & 1m & 91.88 & 89.52    & & 120K & 2m & 98.23 & 96.97 \\

Top-$k$ softmax         & 38K & 20s & 84.29 & 80.62   & & 247K & 1m & 96.60 & 93.60    & & 120K & 2m & 98.54 & 97.14 \\
Top-$k$ 1.5-entmax      & 38K & 20s & 85.44 & 80.28   & & 247K & 1m & 97.88 & 94.92    & & 120K & 2m & 98.22 & 97.37 \\
Top-$k$ sparsemax       & 38K & 20s & 85.44 & 81.54   & & 247K & 1m & 96.76 & 93.32    & & 120K & 2m & 96.46 & 95.72 \\

Select. 1.5-entmax      & 38K & 10s & 85.55 & 80.62   & & 247K & 1m & 97.44 & 94.56    & & 120K & 1m & 98.30 & 97.41 \\
Select. sparsemax       & 38K & 10s & 85.44 & 81.54   & & 247K & 1m & 97.04 & 93.36    & & 120K & 1m & 96.46 & 95.72 \\

Bernoulli               & 38K & 5s & 84.75 & 78.21   & & 247K & 1m & 91.80 & 87.36    & & 120K & 1m & 97.12 & 94.82 \\
HardKuma                & 38K & 5s & 87.50 & 81.76   & & 247K & 1m & 95.36 & 91.20    & & 120K & 1m & 97.38 & 96.05 \\
% 54:40
\bottomrule
\end{tabular}
% }
\end{center}
\caption{Communication results on the validation set and explainer statistics. \# P is the number of trainable parameters, and is $t$ the average training time per epoch.} 

\label{table:communication_statistics_params}

\end{table*}

\begin{table*}[!htb]
\small
\begin{center}
% \resizebox{\textwidth}{!}{%
\begin{tabular}{l cccc c cccc}
\toprule
& \multicolumn{4}{c}{\sc Yelp} & & \multicolumn{4}{c}{\sc SNLI} \\

\cmidrule{2-5} \cmidrule{7-10}

\sc Explainer\,\, & 
\sc \# P & \sc $t$ & CSR & ACC$_L$  & & 
\sc \# P & \sc $t$ & CSR & ACC$_L$  \\

\midrule 

Random                  & 1.8M & 3h & 52.55 & 48.21     & & 560K & 9m  & 31.04 & 33.11 \\
Erasure                 & 1.8M & 4h & 79.63 & 69.59     & & 560K & 10m & 78.72 & 70.60 \\
Top-$k$ gradient        & 1.8M & 3h & 71.81 & 63.59     & & 560K & 10m & 77.55 & 69.41 \\
Top-$k$ softmax         & 1.8M & 3h & 81.49 & 70.67     & & 560K & 9m & 79.10 & 70.95 \\
Top-$k$ 1.5-entmax      & 1.8M & 3h & 82.80 & 71.31     & & 560K & 9m & 80.30 & 73.57 \\
Top-$k$ sparsemax       & 1.8M & 3h & 82.97 & 71.46     & & 560K & 9m & 83.25 & 75.34 \\
Select. 1.5-entmax      & 1.8M & 2h & 82.90 & 70.99     & & 560K & 6m & 77.46 & 71.66 \\
Select. sparsemax       & 1.8M & 2h & 84.67 & 72.25     & & 560K & 6m & 82.33 & 75.11 \\

Bernoulli               & 1.8M & 2h & 84.93 & 66.77     & & 560K & 2m & 75.75 & 68.61 \\
HardKuma                & 1.8M & 2h & 87.43 & 71.57     & & 560K & 3m & 75.10 & 71.10 \\
% 21:10
\bottomrule
\end{tabular}
% }
\end{center}
\caption{Continuation of Table~\ref{table:communication_statistics_params}.} 

\label{table:communication_statistics_params_2}

\end{table*}

\section{Joint $E$ and $L$ setup}

\subsection{Communication}

According to \S\ref{subsec:joint_e_and_l}, in this model we have two set of parameters to train, one for the explainer $E$ and other for the layperson $L$, whereas the classifier is a frozen model that we want to explain. Here, we set $C$ as the RNN with softmax classifier (see \S\ref{table:results_classifiers}). We design $E$ with the same architecture of the RNNs with attention from \S\ref{subsec:rnns_with_attn} but without a final output layer, and $L$ have the same architecture as the laypersons in \S\ref{sec:experiments}. In short, the architecture of $E$  is composed of: (i) embedding layer; (ii) BiLSTM; (iii) attention mechanism. As before, the message is constructed with the words extracted from the attention mechanism. 

We use sparsemax attention during training to ensure end-to-end differentiability, and we recover the top-$k$ attended words during test time. We used $k=5$ for IMDB and $k=4$ for SNLI in all experiments. In order to encourage faithful explanations, we set $h = \frac{1}{L}\sum_i C_{\mathrm{RNN}}(x_i)$ and $\tilde{h} = \frac{1}{L}\sum_i \mathrm{FFN}(E_{\mathrm{RNN}}(x_i))$, where $\mathrm{FFN}$ is a simple feed-forward layer, and $C_{\mathrm{RNN}}(x_i)$ and $E_{\mathrm{RNN}}(x_i)$ are the BiLSTM states from the classifier and the explainer, respectively. In other words, we are approximating the average of the BiLSTM states of $C$ and $E$. We set $\lambda=1$ and $\beta=0.2$ and used the same hyperparameters as in Table~\ref{tab:table_all_hyperparams_comm}. 
The list of stopwords used in our experiments contains 127 English words extracted from NLTK.
%\marcos{say something about testing different values of $\lambda$?}

\subsection{Analysis of $\beta$}\label{sec:analysis_beta}

% \paragraph{Quality of the explanations.} 

A potential problem of this model is for the two agents to agree on a trivial protocol, ensuring a high CSR even with bad quality explanations (e.g. punctuations or stopwords). Besides preventing stopwords to be in the message,\footnote{In practice, we simply set attention scores associated with stopwords to $-\infty$.} we set a different probability $\beta$ of the explainer accessing the predictions of the classifier $\hat{y}$. Intuitively, these strategies should encourage explanations to have higher quality. One way to quantitatively access the quality of the explanations is by aggregating the  relative frequencies of each selected word in the validation set, and calculating its Shannon's entropy. If the entropy is low, then the  explanations have a high number of repetitions and the explainers are focusing on a very small subset of words, denoting a trivial protocol. To check for a reasonable entropy score that resembles a good quality explanation, we investigate the entropy of the other explainers, for which we had confirmed their quality via human evaluation. 
% Table~\ref{tab:entropy_explainers} shows the entropy of each explainer on IMDB and SNLI.

In order to see the impact of $\beta$, we carried an experiment with increasing values of $\beta$ and looked at the CSR, ACC$_L$ and the entropy ($H$) of the generated explanations. Results are shown in Table~\ref{tab:results_joint_betas} for each explainer on IMDB and SNLI.

\begin{table}[!htb]
\small
\begin{center}
\begin{tabular}{ll ccc@{ }c ccc@{ }c}
\toprule
& & \multicolumn{3}{c}{\sc IMDB} & & \multicolumn{3}{c}{\sc SNLI} \\

\cmidrule{3-5} \cmidrule{7-9} 

\sc Clf. & \multirow{1}{*}{\sc Explainer} & $H$ & CSR & ACC$_L$  & & $H$ & CSR & ACC$_L$ \\
% Note: the CSR and ACC of these explainers are different from table 4 for IMDB because here we used k=5 (instead of k=10)
\midrule 
$C$         & Random                  & 9.13 & 59.20 & 58.92 		& & 8.21 & 31.04 & 33.11     \\
$C$         & Erasure                 & 9.40 & 96.32 & 93.48 		& & 9.75 & 78.72 & 70.60     \\
$C$         & Top-$k$ gradient        & 9.49 & 85.84 & 83.72 		& & 9.39 & 77.55 & 69.41     \\
$C$         & Top-$k$ softmax         & 9.38 & 94.44 & 91.84 		& & 9.76 & 78.66 & 71.00     \\
$C_{ent}$   & Top-$k$ 1.5-entmax      & 9.62 & 95.20 & 93.36 		& & 9.54 & 80.30 & 73.57     \\
$C_{sp}$    & Top-$k$ sparsemax       & 9.56 & 95.28 & 92.56 		& & 8.79 & 83.25 & 75.34     \\

$C_{ent}$   & Select. 1.5-entmax      & 10.76 & 97.44 & 94.56 		& & 8.49 & 77.46 & 71.66     \\
$C_{sp}$    & Selec. sparsemax        & 10.41 & 97.04 & 93.36 		& & 8.38 & 82.33 & 75.11     \\
% $C_{bern}$  & Bernoulli               & 9.61 & 91.60 & 87.36 		& & 8.27 & 75.65 & 68.68     \\
% $C_{hk}$    & HardKuma                & 11.54 & 94.20 & 90.80 		& & 9.93 & 71.57 & 68.30     \\
$C_{bern}$  & Bernoulli               & 10.66 & 91.88 & 87.36 		& & 8.27 & 75.75 & 68.61     \\
$C_{hk}$    & HardKuma                & 11.38 & 95.36 & 91.20 		& & 9.93 & 75.10 & 71.10     \\

-           & Human highlights        & -     & -     & - 		    & & 8.72 & 87.97 & 87.97     \\
\midrule 
$C$         & Joint $E$ and $L$ ($\beta=0.0$)        & 6.16 & 93.04 & 90.84   	& & 9.81 & 80.74 & 72.38  \\
$C$         & Joint $E$ and $L$ ($\beta=0.2$)	     & 6.05 & 98.52 & 94.56   	& & 9.81 & 93.44 & 77.20  \\
$C$         & Joint $E$ and $L$ ($\beta=0.5$)	     & 5.63 & 99.68 & 95.64   	& & 9.45 & 95.81 & 77.54  \\
$C$         & Joint $E$ and $L$ ($\beta=1.0$)	     & 3.72 & 99.92 & 95.56   	& & 9.01 & 97.49 & 77.23  \\

\bottomrule
\end{tabular}
\end{center}
\caption{Entropy of the explanations for all explainers on the validation set of IMDB and SNLI. Entropy for human highlights was calculated based on non-neutral examples. 
% $C_\star$ represents a classifier with a 100\% accuracy.
% Explanations for top-$k$ explainers were extracted using $k=5$.
% $S$ denotes the entropy of the explanations.
} 
\label{tab:results_joint_betas}
\end{table}

When $\beta=0$ no information about the label predicted by the classifier is being exposed to the explainer, and as a result we have a model that resembles a combination of selective (during training) and top-$k$ (during test time) sparsemax explainers. This means that the results between these explainers are expected to be very similar in terms of CSR.\footnote{Note that this also depends on the performance of $C$ and $C_{sp}$, which are indeed very similar in this case: 95.64 and 95.44.} Overall, for both datasets, we can see a tradeoff between CSR and entropy $H$ as $\beta$ increases, suggesting that CSR is not able to capture the notion of quality (which was expected due to the subjective nature of an explanation). For IMDB the entropy values were lower than our previous explainers, but for SNLI they were very similar. A potential reason for this is the particularity of the two datasets: IMDB have long documents (280 words on average) with a large set of repetitive words which are not stopwords and are strongly correlated with the labels (e.g. good, ok, bad, etc.); SNLI premises are very short (14 words on average) without a large set of repetitive words.
Finally, due to this tradeoff, we selected $\beta=0.2$ for all of our experiments since it induces a very high CSR with a reasonably good entropy.

\begin{comment}
\paragraph{Degeneration problem.} The problem we call \aspas{trivial protocol} has been diagnosed by \citet{yu2019rethinking} as \aspas{degeneration problem}. Their strategy was to use the notion of comprehensiveness to penalize the predictive power of a classifier trained on the complement set of the selected explanations (i.e. the words that were not selected by the explainer) \marcos{check. Andre can you check this? I'm not sure if they are minimizing or maximizing the entropy of the complement set.}. In contrast, we follow the selective explanation principle of \citet{miller2019explanation} and focus on \textit{sufficient} rationales, adopting a much simpler strategy where we can control the amount of \aspas{degeneration} by carefully setting $\beta$, and selecting the final model by looking at the entropy of the explanations.
\end{comment}

\section{Machine Translation experiments}\label{sec:nmt_setup}

\subsection{Data}

To compare explainers on a more challenging task with large $|\mathcal{Y}|$, we ran an experiment on neural machine translation (NMT), adapting the JoeyNMT framework \citep{kreutzer2019joey}. 
We used the EN$\to$DE IWSLT 2017 dataset \citep{cettolo2017overview}, with the standard splits (Table~\ref{table:datasets}). %We replicated \citet{peters2019sparse} using raw words as input instead of byte-pair encodings. 

\subsection{Classifier} 

We replicated the work of \citet{peters2019sparse} with the exception that we used raw words as input instead of byte-pair encodings. The implementation is based on Joey-NMT \citep{kreutzer2019joey}. 
We employed beam search decoding with beam size of 5, achieving a BLEU score of 20.49, 21.12 and 20.75 for softmax ($C$), 1.5-entmax ($C_{\text{ent}}$) and sparsemax ($C_{\text{sp}}$), respectively. %These values are very close ($<$1.5) to the BLEU scores found in the original work of \citet{peters2019sparse}.$. 
We refer to the work of \citet{peters2019sparse} for more training details. Table~\ref{tab:table_all_hyperparams_nmt} shows the classifier hyperparameters.

% We used the default configuration file from JoeyNMT,\footnote{\url{https://github.com/joeynmt/joeynmt/blob/master/configs/iwslt14_deen_bpe.yaml}} with the following hyperpameters:

\begin{table}[!htb]
    \centering
    \small
    \begin{tabular}{ll}
        \toprule
        \sc Hyperparam. & \sc Value  \\
        \midrule
        Word embeddings size        & 512     \\
        BiRNN hidden size           & 512     \\
        Attention scorer            & \citep{Bahdanau2015} \\
        
        Batch size                  & 32    \\
        Optimizer                   & Adam      \\
        Number of epochs            & 100     \\
        Early stopping patience     & 8 \\
        Learning rate               & 0.001     \\
        % Scheduling                  & Plateau \\
        Decrease factor             & 0.5 \\
        $\ell_2$ regularization     & 0     \\
        
        RNN type                    & LSTM \\
        RNN layers                  & 2 \\
        Dropout                     & 0.3 \\
        Hidden dropout              & 0.3 \\
        % Input feeding               & True \\
        % Init. hidden                & Last \\
        
        Maximum output length & 100 \\
        Beam size & 5 \\
        \bottomrule
    \end{tabular}
    \caption{Classifier hyperparmeters for neural machine translation.}
    \label{tab:table_all_hyperparams_nmt}
\end{table}

\subsection{Communication}

We consider the decision taken by the NMT system when generating the $t^\text{th}$ target word ($y$), given the source sentence $x$ and the previously generated words $y_{1:{t-1}}$. Note that in this example $\mathcal{Y}$ is the entire target vocabulary. 
The message is the concatenation of $k$ source words (ranked by importance, without any word order information) 
% \andre{check} \marcos{what do you mean by order information?} 
% \marcos{Got it. The top k does not include the positional information of the word in the original sentence}
with the prefix $y_{1:{t-1}}$. The layperson must predict the target word given this limited information (see Fig.~\ref{fig:example_mt}). 

\begin{figure}[!htb]
    \centering
    \includegraphics[width=0.6\textwidth]{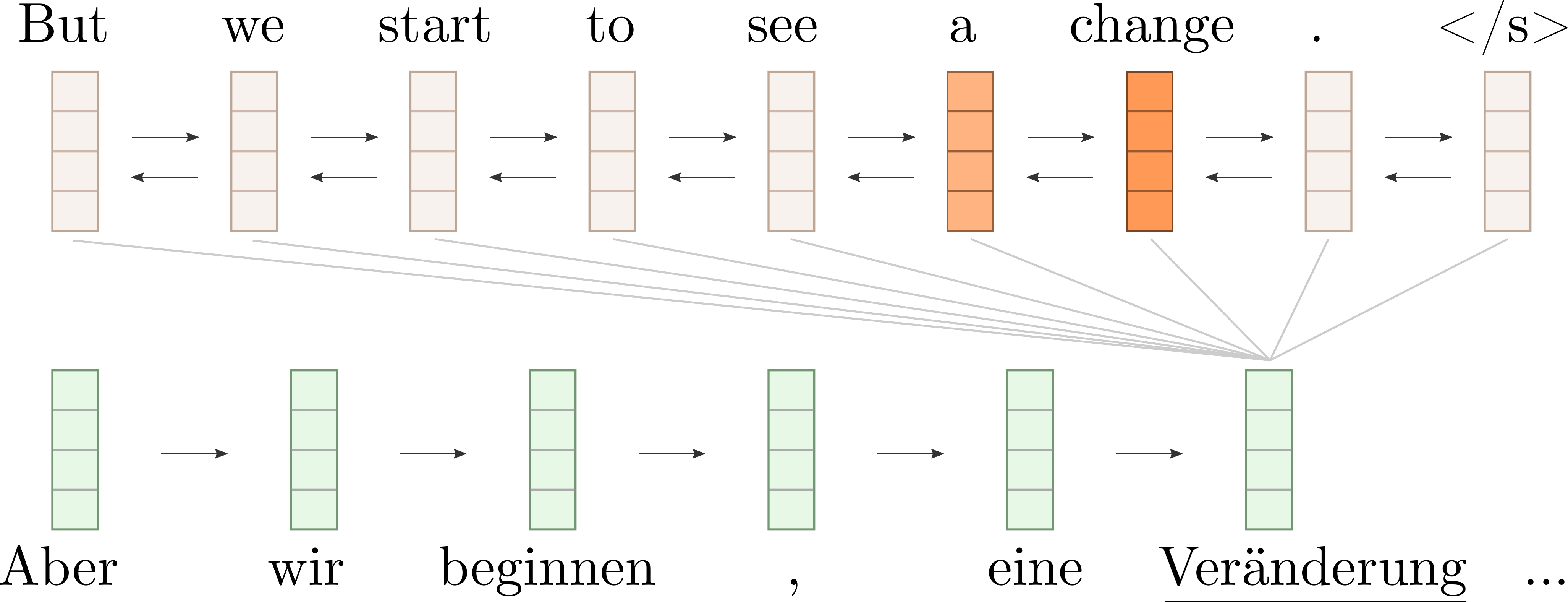}
    \caption{Example of sparse attention for machine translation. When the model is generating the word \aspas{Veränderung}, the source words \aspas{a} and \aspas{change} are treated as explanation and sent as message.}
    % \andre{maybe underline or use a different color for the word Veranderung}
    % Done
    \label{fig:example_mt}
\end{figure}

The layperson is a model that uses an unidirectional LSTM with $256$ hidden units to encode the translation prefix, and a feed-forward layer to encode the concatenation of $k$ source word embeddings (the message) to a vector of $256$ dimensions. 
%This vector is concatenated with the output vector of the translation LSTM for the word being translated at the moment. Finally, 
The two vectors are concatenated and passed to a linear output layer to predict the next  word $\tilde{y} \in \mathcal{Y}$ from the target vocabulary. 
We used 300D-pretrained GloVe embeddings to encode source words (EN), and 300D-pretrained FastText embeddings to encode target words (DE).\footnote{\url{https://dl.fbaipublicfiles.com/fasttext/vectors-crawl/cc.de.300.bin.gz}} 
Table~\ref{tab:table_all_hyperparams_nmt_comm} shows the communication hyperparameters.

\begin{table}[!htb]
    \centering
    \small
    \begin{tabular}{ll}
        \toprule
        \sc Hyperparam. & \sc Value  \\
        \midrule

        Word embeddings size        & 300     \\
        LSTM hidden size            & 256     \\
        Merge LSTM states           & concat \\
        
        Batch size                  & 16    \\
        Number of epochs            & 10    \\
        Early stopping patience     & 5     \\
        Learning rate               & 0.003 \\
        $\ell_2$ regularization     & $10^{-5}$     \\
        \bottomrule
    \end{tabular}
    \caption{Communication hyperparmeters for neural machine translation.}
    \label{tab:table_all_hyperparams_nmt_comm}
\end{table}

\subsection{Results}

Results comparing different filtering methods varying $k$ are shown in Table~\ref{table:results_iwslt}. We show the CSR as we varied $k \in \{0,1,3,5\}$. %, meaning the scores when only the most relevant $k$ source words are shown to the layperson at each translation timestep. 
There are two main findings. First, we see again that {\bf top-$k$ attention outperforms top-$k$ gradient}, in this case with a wider margin. 
Second, we see that  all methods perform better as we increase $k$, albeit we can see a performance degradation of attention-based explainers for $k=5$. An interesting case is when $k=0$, meaning that $L$ has no access to the source sentence, behaving like an unconditioned language model. 
In this case the performance is much worse, indicating that both 
% top-$k$ attention and gradient 
explainers are selecting relevant tokens when $k>0$. %To further emphasize this claim, we trained a model which selects $k=5$ source words at random at each timestep, yielding a CSR of $23.32$, which is very close to the results when $k=0$. %As the findings described in \citep{peters2019sparse}, we also see that 1.5-entmax performs better than softmax and sparsemax.
As we found for IMDB and SNLI, as we increased $k$ we observed a trade-off between $k$ and CSR for IWSLT. Fig.~\ref{fig:k_analysis_nmt} depicts this finding.

% BLEU	    dev	        test	[Beam search decoding with beam size = 5 and alpha = 1.0]
% softmax	20.48768	20.48768	
% sparsemax	24.40572	20.74869	
% entmax15	24.52431	21.12021	

% \andre{include a figure with a real example where the explanation is important to guess the next word?}
% \marcos{I'm on it}

% \begin{itemize}
%     \item qual foi o modelo utilizado para treinar o classifier para machine translation
%     \item quais foram os hiperparametros diferentes do trabalho do Ben
%     \item qual foi o bleu score de cada metodo no conjunto de teste
%     \item como foi implementado o layperson - concat das embeddings (k * emb dim)
%     \item 
% \end{itemize}

% BLEU	    dev	        test
% softmax	20.48768	20.48768
% sparsemax	24.40572	20.74869
% entmax15	24.52431	21.12021

\begin{table}[!htb]
\small
\begin{center}
% \resizebox{\columnwidth}{!}{%
\begin{tabular}{llcccc}
\toprule
%\multirow{2}{*}{\sc Explainer} & \multicolumn{4}{c}{CSR} \\
%\cmidrule{2-5}
\sc Clf. & \sc Explainer & $k=0$ & $k=1$ & $k=3$ & $k=5$ \\
\midrule 
% Uniform    		            & - & - & - & 23.32   	\\
% \midrule
$C$ & Top-$k$ gradient    	& 21.99     & 35.21     & 38.33     & 40.30   	\\
$C$ & Top-$k$ softmax	 	    & 21.99     & \bf 62.58 & 62.82     & 62.64   	\\
$C_{\text{ent}}$ & Top-$k$ 1.5-entmax		& \bf 22.31 & 62.53     & \bf 63.48 & \bf 62.69   	\\
$C_{\text{sp}}$ & Top-$k$ sparsemax	 	& 22.14     & 62.21     & 61.94     & 61.92   	\\
\bottomrule
\end{tabular}
% }
\end{center}
\caption{Results for IWSLT. Reported are CSR scores.
% The classifier $C$ achieves a BLEU score of \andre{?} for this task.
% \marcos{Coloquei isso no texto acima.}
} \label{table:results_iwslt}
\end{table}

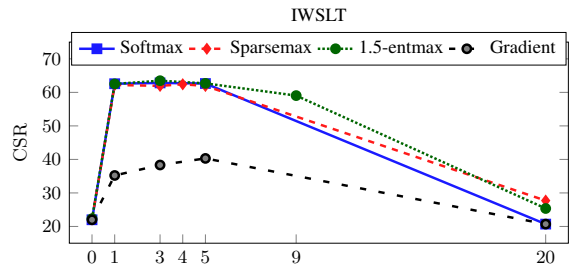
\begin{figure}[!htb]
% for now i'll let the figure this way, if I find time i'll search how to use only one legend
% it is not easy to do this and have two subfigures
    \centering
    
    % IWSLT
    % CSR plot
    \subfloat{{\begin{tikzpicture}[scale=0.78]
\begin{axis}[
    title={IWSLT},
    cycle list name=my black white,
    % xlabel={$k$},
    % ymode=log,
    % xmode=log,
    % log ticks with fixed point,
    % x filter/.code=\pgfmathparse{#1 + 6.90775527898214},
    % axis y line=left,
    ylabel={$\mathrm{CSR}$},
    ymin=15, ymax=75,
    enlarge x limits=0.05,
    legend cell align={left},
    xtick={0,1,3,4,5,9,20},
    % symbolic x coords={0,1,3,4,5,9.38,20},
    ytick={20,30,40,50,60,70},
    % legend pos=north east,
    legend style={at={(0.5,1.05)}, anchor=north,legend columns=-1,font=\small},
    % ymajorgrids=true,
    every axis plot/.append style={very thick},
    % every mark/.append style={mark size=5pt}
    % grid style=dashed,
    % width=15cm,
    tick label style={font=\footnotesize},
    % no markers,
    % max space between ticks={50},
    % log basis x={2},
    % log basis y={2},
    % xtick placement tolerance={25.2},
    % try min ticks log={5},
    % try min ticks={1},
    % minor tick num={1},
    % width=2cm,
    % width=\columnwidth,
    height=5cm,
    font=\small,
]

\addplot
coordinates {
    (0,21.99)    (1,62.58)   (3,62.82)   (5,62.64)   (20,20.70)
};

\addplot
coordinates {
    (0,22.14)    (1,62.21)   (3,61.94)     (4,62.42)     (5,61.92)    (20,27.71)
};

\addplot
coordinates {
    (0,22.31)    (1,62.53)   (3,63.48)   (5,62.69)  (9,59.04)    (20,25.36)
}; 

\addplot
coordinates {
    (0,21.99)    (1,35.21)   (3,38.33)   (5,40.30)   (20,20.70)
};

% Average k	k	csr
% sparsemax	4.53	0.6242
% entmax	9.38	0.5904
		
% maxlen	len	csr
% softmax	154	0.2070
% sparsemax	154	0.2771
% entmax	154	0.2536

\legend{Softmax, Sparsemax, 1.5-entmax, Gradient}

\end{axis}
\end{tikzpicture}}}
    
    \caption{Message sparsity analysis for IWSLT. For SNLI, $k=0$ corresponds to a case where the layperson only sees the translation prefix.
    The rightmost entry is the average length of the examples in the test set, and therefore it represents an explainer that simply pass forward all words to the layperson (i.e. a full bag-of-words). The average $k$ for sparsemax and 1.5-entmax are, respectively: 4.5 and 9.4.
    } \label{fig:k_analysis_nmt}
    
\end{figure}

\section{Human annotation}\label{sec:supp_interface}

% An important aspect of our framework is how it relates to human judgement. 
We had four different human annotators, two for IMDB and two for SNLI. %, who are not the authors of the paper. 
%To prevent potential knowledge bias from the annotators, the name of the explainers was anonimyzed, and to prevent easy comparison across explainers, the documents were displayed in a different order. 
No information was given about the explainers which produced each message, and documents were presented in random order. 
Since in our experiments we define the message as being a bag-of-words, which does not encode order information, the explanations (i.e. the selected words) were shuffled and displayed as a cloud of words. The annotators were asked to predict the label of each document, when seeing only these explanations. For SNLI, we show the entire hypothesis as raw text and the premise as a cloud of words. We selected top-$k$ explainers with $k=5$ for IMDB and $k=4$ for SNLI. Figure~\ref{fig:interface_attn} shows a snapshot of the annotation interface used for the experiments described in \S\ref{sec:human_evaluation}.

\begin{figure*}[!htb]
    \centering
    \centering
    \subfloat[IMDB]{\includegraphics[width=0.37\textwidth]{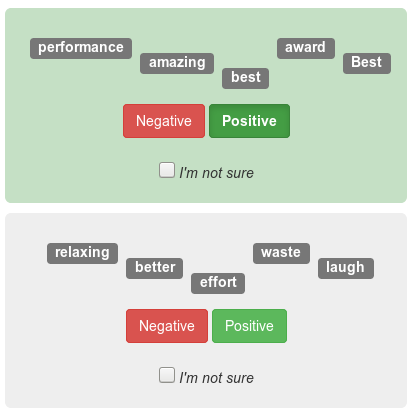}}
    \quad
    \subfloat[SNLI]{\includegraphics[width=0.37\textwidth]{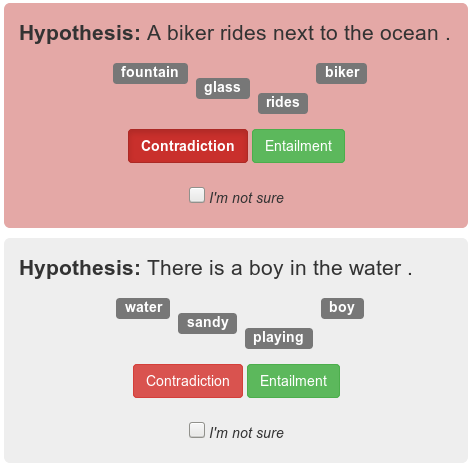}}
    \caption{Snapshot of the annotation interface.}
    \label{fig:interface_attn}
\end{figure*}

By directly looking at the explanations, we observed that some of them are very ambiguous with respect to the true label, so we decided to include a checkbox to be marked in case the annotator was not sure by his/her decision. The unsure checkbox also helps to capture the notion of sufficiency, that is, if the explanations are sufficient for a human predict some label. A similar approach was employed by \citet{yu2019rethinking} using a two-stage annotation method, explicitly asking the human annotator if the rationale was sufficient for his/her decision. 
Furthermore, we calculated the agreement between explainers using the Cohen's kappa coefficient and the relative observed agreement ratio (or accuracy, $p_o$). Table~\ref{tab:table_agreement_human_eval} shows statistics for the unsure checkbox and agreement between annotators.

\begin{table}[!htb]
\small
\begin{center}

\begin{tabular}{ll ccccc c ccccc}
\toprule
\multirow{2}{*}{\sc Clf.} &
\multirow{2}{*}{\sc Explainer} & \multicolumn{5}{c}{IMDB} & & \multicolumn{5}{c}{SNLI} \\
\cmidrule{3-7}
\cmidrule{9-13}
        & & $u$ & $p_o$ & $\kappa$ & CSR$_H$ & ACC$_H$         & & $u$ & $p_o$ & $\kappa$ & CSR$_H$ & ACC$_H$ \\    
\midrule 
$C$ & Erasure 	                        & 0.05 & 0.92 & 0.83 & 89.25 & 86.25 		& & 0.25 & 0.83 & 0.66 & 72.50 & 83.50   \\
$C$ & Top-$k$ gradient 					& 0.17 & 0.76 & 0.51 & 73.50 & 73.00 	    & &	0.32 & 0.80 & 0.59 & 65.75 & 76.75    \\
$C$ & Top-$k$ softmax 					& 0.23 & 0.91 & 0.81 & 89.25 & 88.25 	    & &	0.25 & 0.78 & 0.55 & 72.00 & 82.75   \\
$C_{\text{ent}}$ & Top-$k$ 1.5-entmax 	& 0.09 & 0.91 & 0.81 & 89.25 & 85.75  	    & &	0.29 & 0.82 & 0.64 & 70.00 & 80.50   \\
$C_{\text{sp}}$ & Top-$k$ sparsemax 	& 0.09 & 0.88 & 0.76 & 89.00 & 87.50  	    & &	0.38 & 0.80 & 0.59 & 68.25 & 80.25   \\
\midrule
$C_{\text{ent}}$ & Selec. 1.5-entmax 	& 0.13 & 0.80 & 0.60 & 86.50 & 84.00 		& & 0.21 & 0.84 & 0.67 & 75.25 & 87.00   \\
$C_{\text{sp}}$ & Selec. sparsemax 		& 0.10 & 0.89 & 0.77 & 87.75 & 86.75 		& & 0.35 & 0.83 & 0.66 & 72.25 & 85.00   \\
% $C_{\text{bern}}$ & Bernoulli 			& 0.42 & 0.66 & 0.30 & 70.75 & 70.75 		& & 0.24 & 0.85 & 0.69 & 74.50 & 86.75  \\
% $C_{\text{hk}}$ & HardKuma 				& 0.58 & 0.60 & 0.17 & 67.00 & 68.00 		& &	0.18 & 0.86 & 0.72 & 79.25 & 87.50 	\\
$C_{\text{bern}}$ & Bernoulli 			& 0.25 & 0.72 & 0.43 & 79.00 & 75.00 		& & 0.24 & 0.85 & 0.69 & 74.50 & 86.75  \\
$C_{\text{hk}}$ & HardKuma 				& 0.17 & 0.81 & 0.61 & 83.75 & 80.75 		& &	0.18 & 0.86 & 0.72 & 79.25 & 87.50 	\\
\midrule
$C$ & Joint $E$ and $L$ 				& 0.12 & 0.96 & 0.91 & 96.75 & 89.25 	    & &	0.65 & 0.71 & 0.44 & 58.00 & 70.00 	\\
- & Human highlights					& - & -   & -  & -  & -      		        & &	0.34 & 0.88 & 0.74 & 83.25 & 83.25 	\\
\midrule
% \multicolumn{2}{l}{\bf Average}         & \bf 0.20 & \bf 0.83 & \bf 0.65 & - & -    & & \bf 0.31 & \bf 0.82 & \bf 0.63 & - & - \\
\multicolumn{2}{l}{\bf Average}         & \bf 0.14 & \bf 0.85 & \bf 0.70 & - & -    & & \bf 0.31 & \bf 0.82 & \bf 0.63 & - & - \\
\bottomrule
\end{tabular}
\end{center}
\caption{Results for human evaluation. $\kappa$ is the Cohen's kappa coefficient, $p_o$ is the relative observed agreement, and $u$ represents the average of the portion of examples where annotators were unsure about their decisions.} 
\label{tab:table_agreement_human_eval}
\end{table}

% >>> np.mean(x1==x2)
% 0.845
% >>> cohen_kappa_score(x1,x2)
% 0.6854707792207793

\section{Examples of explanations}\label{sec:supp_explanations}

Tables \ref{table:word_overlap_imdb} and \ref{table:word_overlap_snli} show the average word overlap between explainers' messages ($m$) for IMDB and SNLI. Looking at the statistics we observed that, in general, top-$k$ attention-based classifiers produce similar explanations among themselves, and the erasure explainer produces messages similar to top-$k$ softmax. Major differences are observed for top-$k$ gradient and rationalizers, while selective attention produces, by definition, more words than top-$k$ attention (i.e. $m_{\text{top-k}} \subseteq m_{\text{selective}}$). It is worth noticing that although explainers with similar messages are expected to have a similar CSR (e.g. top-$k$ attention and erasure), including/excluding a single word in the explanation might impact the layperson decision, as we can see in the next examples. 
Tables \ref{table:examples_explanations_imdb_csrs} and \ref{table:examples_explanations_imdb_csrs2} show the output of erasure, gradient, attention, and joint explainers for IMDB, along with the prediction made by the classifier ($y_C$) and the layperson ($y_L$). In Tables \ref{table:examples_explanations_snli_csrs} and \ref{table:examples_explanations_snli_csrs2} we also include the human highlights explainer for SNLI. 
% In Table~\ref{table:examples_explanations_imdb_csrs} we show outputs for IMDB cases when the classifier and the layperson predicted a different label (decreasing CSR). Table~\ref{table:examples_explanations_snli_csrs} shows the same but for SNLI.

% \marcos{shuffle explanations in these tables? - except bernoulli and hardkuma? dont know}

\begin{table}[!htb]
\scriptsize
\begin{center}

\begin{tabular}{@{}lllllllllll@{}}
\toprule

& Erasure 
& \begin{tabular}[c]{@{}l@{}}Top-$k$\\gradient\end{tabular} 
& \begin{tabular}[c]{@{}l@{}}Top-$k$\\softmax\end{tabular} 
& \begin{tabular}[c]{@{}l@{}}Top-$k$\\1.5-entmax\end{tabular} 
& \begin{tabular}[c]{@{}l@{}}Top-$k$\\sparsemax\end{tabular} 
& \begin{tabular}[c]{@{}l@{}}Selec.\\1.5-entmax\end{tabular} 
& \begin{tabular}[c]{@{}l@{}}Selec.\\sparsemax\end{tabular} 
& Bernoulli 
& HardKuma 
& Joint $E$ and $L$ \\
\midrule 
Erasure             & \emph{1.00}  & 0.34  & 0.85  & 0.56  & 0.55  & 0.20  & 0.37  & 0.14  & 0.23  & 0.20 \\ 
Top-$k$ gradient    & 0.34	& \emph{1.00}	& 0.35	& 0.30	& 0.30	& 0.16	& 0.26	& 0.11	& 0.18	& 0.11 \\
Top-$k$ softmax     & 0.85	& 0.35	& \emph{1.00}	& 0.57	& 0.55	& 0.20	& 0.37	& 0.14	& 0.24	& 0.20 \\
Top-$k$ 1.5-entmax  & 0.56	& 0.30	& 0.57	& \emph{1.00}	& 0.61	& 0.21	& 0.39	& 0.12	& 0.24	& 0.19 \\
Top-$k$ sparsemax   & 0.55	& 0.30	& 0.55	& 0.61	& \emph{1.00}	& 0.20	& 0.43	& 0.13	& 0.24	& 0.20 \\
Selec. 1.5-entmax   & 0.20	& 0.16	& 0.20	& 0.21	& 0.20	& \emph{1.00}	& 0.45	& 0.24	& 0.44	& 0.08 \\
Selec. sparsemax    & 0.37	& 0.26	& 0.37	& 0.39	& 0.43	& 0.45	& \emph{1.00}	& 0.21	& 0.41	& 0.13 \\
Bernoulli           & 0.14	& 0.11	& 0.14	& 0.12	& 0.13	& 0.24	& 0.21	& \emph{1.00}	& 0.28	& 0.06 \\
HardKuma            & 0.23	& 0.18	& 0.24	& 0.24	& 0.24	& 0.44	& 0.41	& 0.28	& \emph{1.00}	& 0.08 \\
Joint $E$ and $L$   & 0.20	& 0.11	& 0.20	& 0.19	& 0.20	& 0.08	& 0.13	& 0.06	& 0.08	& \emph{1.00} \\

\bottomrule
\end{tabular}
\end{center}
\caption{Average word overlap (\%) between explainers for IMDB.} 
\label{table:word_overlap_imdb}
\end{table}
\begin{table}[!htb]
\scriptsize
\begin{center}

\begin{tabular}{@{}lllllllllll@{}}
\toprule

& Erasure 
& \begin{tabular}[c]{@{}l@{}}Top-$k$\\gradient\end{tabular} 
& \begin{tabular}[c]{@{}l@{}}Top-$k$\\softmax\end{tabular} 
& \begin{tabular}[c]{@{}l@{}}Top-$k$\\1.5-entmax\end{tabular} 
& \begin{tabular}[c]{@{}l@{}}Top-$k$\\sparsemax\end{tabular} 
& \begin{tabular}[c]{@{}l@{}}Selec.\\1.5-entmax\end{tabular} 
& \begin{tabular}[c]{@{}l@{}}Selec.\\sparsemax\end{tabular} 
& Bernoulli 
& HardKuma 
& Joint $E$ and $L$ \\
\midrule 
Erasure             & \emph{1.00} & 0.38 & 0.77 & 0.55 & 0.41 & 0.35 & 0.37 & 0.32 & 0.49 & 0.38 \\
Top-$k$ gradient    & 0.38 & \emph{1.00} & 0.40 & 0.36 & 0.31 & 0.34 & 0.33 & 0.32 & 0.35 & 0.26 \\
Top-$k$ softmax     & 0.77 & 0.40 & \emph{1.00} & 0.56 & 0.41 & 0.36 & 0.37 & 0.32 & 0.49 & 0.38 \\
Top-$k$ 1.5-entmax  & 0.55 & 0.36 & 0.56 & \emph{1.00} & 0.46 & 0.36 & 0.42 & 0.32 & 0.46 & 0.34 \\
Top-$k$ sparsemax   & 0.41 & 0.31 & 0.41 & 0.46 & \emph{1.00} & 0.35 & 0.54 & 0.32 & 0.38 & 0.29 \\
Selec. 1.5-entmax   & 0.36 & 0.34 & 0.36 & 0.36 & 0.35 & \emph{1.00} & 0.64 & 0.88 & 0.48 & 0.26 \\
Selec. sparsemax    & 0.37 & 0.33 & 0.37 & 0.42 & 0.54 & 0.64 & \emph{1.00} & 0.60 & 0.45 & 0.26 \\
Bernoulli           & 0.32 & 0.32 & 0.32 & 0.32 & 0.32 & 0.88 & 0.60 & \emph{1.00} & 0.46 & 0.24 \\
HardKuma            & 0.49 & 0.35 & 0.49 & 0.46 & 0.38 & 0.48 & 0.45 & 0.46 & \emph{1.00} & 0.38 \\
Joint $E$ and $L$   & 0.38 & 0.26 & 0.38 & 0.34 & 0.29 & 0.26 & 0.26 & 0.24 & 0.38 & \emph{1.00} \\

\bottomrule
\end{tabular}
\end{center}
\caption{Average word overlap (\%) between explainers for SNLI.} 
\label{table:word_overlap_snli}
\end{table}

\begin{table*}[!htb]
\scriptsize
\begin{center}
% \resizebox{\textwidth}{!}{%
\begin{tabular}{lllp{11cm}}

\toprule

\multicolumn{4}{p{16cm}}{\textbf{(\textcolor{green!40!black}{positive})} Mardi Gras : Made in china is an excellent movie that depicts how two cultures have much in common but , are not even aware of the influence each society has on one another . David Redmon open your eyes and allows you to see how the workers in china manufactures beads that cost little to nothing and are sold in America for up to 20 dollars . When Redmon questions Americans about where these beads come from they had no clue and seemed dumb founded . When he told them that they are made in China for less then nothing with horrible pay and unacceptable working conditions , Americans seemed sad , hurt , and a little remorseful but didn ' t really seem that they would stop purchasing the beads after finding out the truth . When Redmon questioned the workers in china they did not know that Americans were wearing them over their necks and paid so much for these beads . The workers laughed at what the purpose was behind beads and couldn ' t believe it . This movie is a great film that gives us something to think about in other countries besides our own . $<$ br $>$ $<$ br $>$ M . Pitts}
\\
\midrule

\sc Explainer & $y_C$ & $y_L$  &  \sc Explanation \\

\midrule

Erasure 	        & \textcolor{green!40!black}{pos} & \textcolor{green!40!black}{pos} & 	 excellent great film besides hurt
\\
Top-$k$ gradient 	& \textcolor{green!40!black}{pos} & \textcolor{orange!70!black}{neg} & 	 hurt horrible a excellent couldn
\\
Top-$k$ softmax 	& \textcolor{green!40!black}{pos} & \textcolor{green!40!black}{pos} & 	 excellent great film movie besides
\\
Top-$k$ 1.5-entmax 	& \textcolor{green!40!black}{pos} & \textcolor{green!40!black}{pos} & 	 great excellent couldn that besides
\\
Top-$k$ sparsemax 	& \textcolor{green!40!black}{pos} & \textcolor{green!40!black}{pos} & 	 excellent great couldn gives besides
\\
Select. entmax15 	& \textcolor{green!40!black}{pos} & \textcolor{green!40!black}{pos} & 	 great excellent couldn that besides hurt didn that horrible is china Pitts gives us Redmon stop is not for t
\\
Select. sparsemax 	& \textcolor{green!40!black}{pos} & \textcolor{green!40!black}{pos} & 	 excellent great couldn gives besides china hurt that is
\\
Bernoulli 	        & \textcolor{green!40!black}{pos} & \textcolor{orange!70!black}{neg} & 	 an excellent movie another dumb horrible unacceptable sad remorseful movie great br br Pitts
\\
HardKuma         	& \textcolor{green!40!black}{pos} & \textcolor{green!40!black}{pos} & 	 excellent movie depicts America dumb horrible a great gives us besides our Pitts
\\
Joint $E$ and $L$ 	& \textcolor{green!40!black}{pos} & \textcolor{green!40!black}{pos} & 	 great excellent
\\

\midrule
\\

\multicolumn{4}{p{16cm}}{\textbf{(\textcolor{orange!70!black}{negative})} I don ' t remember " Barnaby Jones " being no more than a very bland , standard detective show in which , as per any Quinn Martin show , Act I was the murder , Act II was the lead character figuring out the murder , Act III was the plot twist ( another character murdered ), Act IV was the resolution and the Epilogue was Betty ( Lee Meriwether ) asking her father - in - law Barnaby Jones ( Buddy Ebsen ) how he figured out the crime and then someone saying something witty at the end of the show . $<$ br $>$ $<$ br $>$ One thing I do remember was the late , great composer Jerry Goldsmith ' s excellent theme song . Strangely , the opening credit sequence made me want to see the show off and on for the seven seasons the show was on the air . I will also admit that it was nice to see Ebsen in a role other than Jed Clampett despite Ebsen being badly miscast . I just wished the show was more entertaining than when I first remembered it . $<$ br $>$ $<$ br $>$ Update ( 1 / 11 / 2009 ): I watched an interview with composer Jerry Goldsmith on YouTube through their Archive of American Television channel . Let ' s just say that I was more kind than Goldsmith about the show " Barnaby Jones ."
} \\

\midrule

\sc Explainer & $y_C$ & $y_L$  &  \sc Explanation \\

\midrule

Erasure 	        & \textcolor{orange!70!black}{neg} & \textcolor{green!40!black}{pos} & 	 wished excellent remembered miscast Strangely
\\
Top-$k$ gradient 	& \textcolor{orange!70!black}{neg} & \textcolor{orange!70!black}{neg} & 	 miscast excellent remembered it badly
\\
Top-$k$ softmax 	& \textcolor{orange!70!black}{neg} & \textcolor{green!40!black}{pos} & 	 wished excellent remembered miscast figuring
\\
Top-$k$ 1.5-entmax 	& \textcolor{orange!70!black}{neg} & \textcolor{orange!70!black}{neg} & 	 wished remembered Strangely miscast excellent
\\
Top-$k$ sparsemax 	& \textcolor{orange!70!black}{neg} & \textcolor{orange!70!black}{neg} & 	 Strangely miscast wished badly excellent
\\
Select. entmax15 	& \textcolor{orange!70!black}{neg} & \textcolor{orange!70!black}{neg} & 	 wished remembered Strangely miscast excellent admit bland no character figuring say badly figured credit , the $<$ the witty want just thing $<$
\\
Select. sparsemax 	& \textcolor{orange!70!black}{neg} & \textcolor{orange!70!black}{neg} & 	 Strangely miscast wished badly excellent remembered bland
\\
Bernoulli 	& \textcolor{orange!70!black}{neg} & \textcolor{orange!70!black}{neg} & 	 very bland , lead character plot character Epilogue witty show br late composer excellent theme song Strangely seasons nice badly miscast entertaining remembered br ( 1 / composer American Television
\\
HardKuma 	& \textcolor{orange!70!black}{neg} & \textcolor{orange!70!black}{neg} & 	 bland figuring saying excellent Strangely credit admit miscast wished remembered ( 1 11
\\
Joint $E$ and $L$ 	& \textcolor{orange!70!black}{neg} & \textcolor{orange!70!black}{neg} & 	 bland badly something
\\

\midrule
\\

\multicolumn{4}{p{16cm}}{\textbf{(\textcolor{green!40!black}{positive})} 
Yes ... I ' m going with the 1 - 0 on this and here ' s why . In the last few years , I have watched quite a few comedies and only left with a few mild laughs and a couple video rental late fees because the movies were that easy to forget . Then I stumble upon " Nothing ". Looked interesting , wasn ' t expecting much though . I was wrong . This was probably one of the funniest movies I have ever had the chance to watch . Dave and Andrew make a great comedic pair and the humor was catchy enough to remember , but not over complex to the point of missing the joke . I don ' t want to remark on any of the actual scenes , because I do feel this is a movie worth seeing for once . With more and more pointless concepts coming into movies ( you know , like killer military jets and " fresh " remakes that are ruining old classics ), This movie will make you happy to say it ' s OK to laugh at " Nothing ".
} \\

\midrule

\sc Explainer & $y_C$ & $y_L$  &  \sc Explanation \\

\midrule

Erasure 	        & \textcolor{green!40!black}{pos} & \textcolor{green!40!black}{pos} & 	 funniest worth great wrong pointless
\\
Top-$k$ gradient 	& \textcolor{green!40!black}{pos} & \textcolor{green!40!black}{pos} & 	 comedic funniest OK worth joke
\\
Top-$k$ softmax 	& \textcolor{green!40!black}{pos} & \textcolor{green!40!black}{pos} & 	 funniest worth great wrong pointless
\\
Top-$k$ 1.5-entmax 	& \textcolor{green!40!black}{pos} & \textcolor{green!40!black}{pos} & 	 funniest great wrong worth not
\\
Top-$k$ sparsemax 	& \textcolor{green!40!black}{pos} & \textcolor{green!40!black}{pos} & 	 funniest worth great catchy wrong
\\
Select. entmax15 	& \textcolor{green!40!black}{pos} & \textcolor{orange!70!black}{neg} & 	 funniest great wrong worth not catchy do probably pointless easy feel ruining movie OK joke ever Yes seeing stumble comedic mild don wasn enough ), forget because 0 for
\\
Select. sparsemax 	& \textcolor{green!40!black}{pos} & \textcolor{orange!70!black}{neg} & 	 funniest worth great catchy wrong ruining 0 feel easy OK not pointless
\\
Bernoulli 	        & \textcolor{orange!70!black}{neg} & \textcolor{orange!70!black}{neg} & 	 - few comedies few mild laughs couple movies stumble interesting wrong probably funniest movies Dave great comedic humor catchy joke scenes movie pointless movies fresh remakes ruining movie Nothing ".
\\
HardKuma 	        & \textcolor{orange!70!black}{neg} & \textcolor{orange!70!black}{neg} & 	 0 stumble wrong probably one funniest great catchy not joke a movie worth seeing pointless ruining OK Nothing
\\
Joint $E$ and $L$ 	& \textcolor{green!40!black}{pos} & \textcolor{orange!70!black}{neg} & 	 funniest pointless worth
\\

\midrule
\\

\multicolumn{4}{p{16cm}}{\textbf{(\textcolor{orange!70!black}{negative})} 
I ' m not to keen on The Pallbearer , it ' s not too bad , but just very slow at the times . As the movie goes on , it gets a little more interesting , but nothing brilliant . I really like David Schwimmer and I think he ' s good here . I ' m not a massive Gwyneth Paltrow fan , but I don ' t mind her sometimes and she ' s okay here . The Pallbearer is not a highly recommended movie , but if you like the leads then you might enjoy it .
} \\

\midrule

\sc Explainer & $y_C$ & $y_L$  &  \sc Explanation \\

\midrule

Erasure 	        & \textcolor{orange!70!black}{neg} & \textcolor{green!40!black}{pos} & 	 brilliant slow recommended nothing good
\\
Top-$k$ gradient 	& \textcolor{orange!70!black}{neg} & \textcolor{green!40!black}{pos} & 	 not nothing recommended slow brilliant
\\
Top-$k$ softmax 	& \textcolor{orange!70!black}{neg} & \textcolor{green!40!black}{pos} & 	 brilliant slow nothing recommended good
\\
Top-$k$ 1.5-entmax 	& \textcolor{orange!70!black}{neg} & \textcolor{orange!70!black}{neg} & 	 slow brilliant nothing not recommended
\\
Top-$k$ sparsemax 	& \textcolor{green!40!black}{pos} & \textcolor{green!40!black}{pos} & 	 slow brilliant nothing recommended good
\\
Select. entmax15 	& \textcolor{orange!70!black}{neg} & \textcolor{green!40!black}{pos} & 	 slow brilliant nothing not recommended good enjoy highly very if you goes don okay , little it bad gets really
\\
Select. sparsemax 	& \textcolor{green!40!black}{pos} & \textcolor{green!40!black}{pos} & 	 slow brilliant nothing recommended good enjoy very bad highly
\\
Bernoulli 	        & \textcolor{orange!70!black}{neg} & \textcolor{orange!70!black}{neg} & 	 Pallbearer , too bad slow times movie , brilliant good massive okay Pallbearer highly movie enjoy
\\
HardKuma 	        & \textcolor{orange!70!black}{neg} & \textcolor{green!40!black}{pos} & 	 slow nothing brilliant good okay highly recommended might enjoy
\\
Joint $E$ and $L$ 	& \textcolor{orange!70!black}{neg} & \textcolor{orange!70!black}{neg} & 	 nothing bad slow okay highly
\\

\bottomrule
\end{tabular}
% }
\end{center}
\caption{Examples of extracted explanations for IMDB.} \label{table:examples_explanations_imdb_csrs}
\end{table*}

%%%%%%%%%%%%%%%%%%%%%%%%%%%%%%%%%%%%%%%%%%%%%%%%%%%%%%%%%%%%%%%%%%%%%%%%%%%%%%%%%%%%%%%%%%%%%%%%
%%%%%%%%%%%%%%%%%%%%%%%%%%%%%%%%%%%%%%%%%%%%%%%%%%%%%%%%%%%%%%%%%%%%%%%%%%%%%%%%%%%%%%%%%%%%%%%%
%%%%%%%%%%%%%%%%%%%%%%%%%%%%%%%%%%%%%%%%%%%%%%%%%%%%%%%%%%%%%%%%%%%%%%%%%%%%%%%%%%%%%%%%%%%%%%%%
%%%%%%%%%%%%%%%%%%%%%%%%%%%%%%%%%%%%%%%%%%%%%%%%%%%%%%%%%%%%%%%%%%%%%%%%%%%%%%%%%%%%%%%%%%%%%%%%
%%%%%%%%%%%%%%%%%%%%%%%%%%%%%%%%%%%%%%%%%%%%%%%%%%%%%%%%%%%%%%%%%%%%%%%%%%%%%%%%%%%%%%%%%%%%%%%%
%%%%%%%%%%%%%%%%%%%%%%%%%%%%%%%%%%%%%%%%%%%%%%%%%%%%%%%%%%%%%%%%%%%%%%%%%%%%%%%%%%%%%%%%%%%%%%%%
%%%%%%%%%%%%%%%%%%%%%%%%%%%%%%%%%%%%%%%%%%%%%%%%%%%%%%%%%%%%%%%%%%%%%%%%%%%%%%%%%%%%%%%%%%%%%%%%
%%%%%%%%%%%%%%%%%%%%%%%%%%%%%%%%%%%%%%%%%%%%%%%%%%%%%%%%%%%%%%%%%%%%%%%%%%%%%%%%%%%%%%%%%%%%%%%%
%%%%%%%%%%%%%%%%%%%%%%%%%%%%%%%%%%%%%%%%%%%%%%%%%%%%%%%%%%%%%%%%%%%%%%%%%%%%%%%%%%%%%%%%%%%%%%%%
%%%%%%%%%%%%%%%%%%%%%%%%%%%%%%%%%%%%%%%%%%%%%%%%%%%%%%%%%%%%%%%%%%%%%%%%%%%%%%%%%%%%%%%%%%%%%%%%
%%%%%%%%%%%%%%%%%%%%%%%%%%%%%%%%%%%%%%%%%%%%%%%%%%%%%%%%%%%%%%%%%%%%%%%%%%%%%%%%%%%%%%%%%%%%%%%%
%%%%%%%%%%%%%%%%%%%%%%%%%%%%%%%%%%%%%%%%%%%%%%%%%%%%%%%%%%%%%%%%%%%%%%%%%%%%%%%%%%%%%%%%%%%%%%%%

\begin{table*}[!htb]
\scriptsize
\begin{center}
% \resizebox{\textwidth}{!}{%
\begin{tabular}{lllp{12cm}}

\toprule

\multicolumn{4}{p{16cm}}{\textbf{(\textcolor{green!40!black}{positive})} Ok , when I rented this several years ago I had the worst expectations . Yes , the acting isn ' t great , and the picture itself looks dated , but as I sat there , a strange thing happened . I started to like it . The action is great and there are few scenes that make you jump . Brion James , maybe one of the greatest B - grade actors next to Bruce Campbell , is great as always . The story isn ' t bad either . Now I wouldn ' t rush out and buy it , but you won ' t waste your time at least watching this good b - grade post apocalyptic western .}
\\
\midrule

\sc Explainer & $y_C$ & $y_L$  &  \sc Explanation \\

\midrule

Erasure 	& \textcolor{green!40!black}{pos} & \textcolor{green!40!black}{pos} & 	 good great great grade waste
\\
Top-$k$ gradient 	& \textcolor{green!40!black}{pos} & \textcolor{orange!70!black}{neg} & 	 waste worst greatest grade t
\\
Top-$k$ softmax 	& \textcolor{green!40!black}{pos} & \textcolor{orange!70!black}{neg} & 	 good great great worst grade
\\
Top-$k$ 1.5-entmax 	& \textcolor{green!40!black}{pos} & \textcolor{green!40!black}{pos} & 	 great waste great good greatest
\\
Top-$k$ sparsemax 	& \textcolor{green!40!black}{pos} & \textcolor{orange!70!black}{neg} & 	 great waste great good grade
\\
Select. entmax15 	& \textcolor{green!40!black}{pos} & \textcolor{green!40!black}{pos} & 	 great waste great good greatest great always Ok apocalyptic Yes make buy t grade isn worst but wouldn strange is
\\
Select. sparsemax 	& \textcolor{green!40!black}{pos} & \textcolor{orange!70!black}{neg} & 	 great waste great good grade greatest your worst Yes Ok
\\
Bernoulli 	& \textcolor{orange!70!black}{neg} & \textcolor{orange!70!black}{neg} & 	 worst , acting , looks strange great scenes greatest actors great story bad , waste watching good apocalyptic western
\\
HardKuma 	& \textcolor{green!40!black}{pos} & \textcolor{orange!70!black}{neg} & 	 worst great great always waste good apocalyptic
\\
Joint $E$ and $L$ 	& \textcolor{green!40!black}{pos} & \textcolor{orange!70!black}{neg} & 	 great worst
\\

\midrule
\\

\multicolumn{4}{p{16cm}}{\textbf{(\textcolor{orange!70!black}{negative})} I have read each and every one of Baroness Orczy ' s Scarlet Pimpernel books . Counting this one , I have seen 3 pimpernel movies . The one with Jane Seymour and Anthony Andrews i preferred greatly to this . It goes out of its way for violence and action , occasionally completely violating the spirit of the book . I don ' t expect movies to stick directly to plots , i gave up being that idealistic long ago , but if an excellent movie of a book has already been made , don ' t remake it with a tv movie that includes excellent actors and nice costumes , but a barely decent script . Sticking with the 80 ' s version .... Rahne
} \\

\midrule

\sc Explainer & $y_C$ & $y_L$  &  \sc Explanation \\

\midrule

Erasure 	& \textcolor{orange!70!black}{neg} & \textcolor{green!40!black}{pos} & 	 excellent excellent script barely decent
\\
Top-$k$ gradient 	& \textcolor{orange!70!black}{neg} & \textcolor{orange!70!black}{neg} & 	 barely decent script if but
\\
Top-$k$ softmax 	& \textcolor{orange!70!black}{neg} & \textcolor{green!40!black}{pos} & 	 excellent excellent script decent barely
\\
Top-$k$ 1.5-entmax 	& \textcolor{orange!70!black}{neg} & \textcolor{green!40!black}{pos} & 	 barely excellent excellent have Sticking
\\
Top-$k$ sparsemax 	& \textcolor{orange!70!black}{neg} & \textcolor{green!40!black}{pos} & 	 excellent excellent barely pimpernel decent
\\
Select. entmax15 	& \textcolor{orange!70!black}{neg} & \textcolor{green!40!black}{pos} & 	 barely excellent excellent have Sticking preferred decent . It don to t script way if costumes Counting pimpernel Rahne , nice greatly t have
\\
Select. sparsemax 	& \textcolor{orange!70!black}{neg} & \textcolor{green!40!black}{pos} & 	 excellent excellent barely pimpernel decent preferred nice t Sticking It
\\
Bernoulli 	& \textcolor{green!40!black}{pos} & \textcolor{green!40!black}{pos} & 	 Baroness Orczy pimpernel movies greatly occasionally movies plots excellent movie tv excellent actors nice costumes barely decent script ....
\\
HardKuma 	& \textcolor{orange!70!black}{neg} & \textcolor{green!40!black}{pos} & 	 have pimpernel preferred way excellent excellent barely decent Sticking Rahne
\\
Joint $E$ and $L$ 	& \textcolor{orange!70!black}{neg} & \textcolor{orange!70!black}{neg} & 	 barely expect decent preferred completely
\\

\midrule
\\

\multicolumn{4}{p{16cm}}{\textbf{(\textcolor{orange!70!black}{negative})} While I agree that this was the most horrendous movie ever made , I am proud to say I own a copy simply because myself and a bunch of my friends were extras ( mostly in the dance club scenes , but a few others as well . This movie had potential with Bolo and the director of Enter the Dragon signed on , but as someone who was on set most every day I can tell you that Robert Clouse was an old and confused individual , at least during the making of this movie . It was a wonder he could find his way to the set everyday . I would also like to think that this might have been a better movie if a lot of it had not been destroyed in a fire at Morning Calm studios . I can ' t say that it would have been for sure , but it would be nice to think so . I was actually surprised that it was ever released , and that someone like Bolo would attach his name to it without a fight . Oh well . Also look at the extras for pro wrestler Scott Levy , AKA Raven . He was a wrestler in Portland at the time ... nice guy , very smart .
} \\

\midrule

\sc Explainer & $y_C$ & $y_L$  &  \sc Explanation \\

\midrule

Erasure 	& \textcolor{orange!70!black}{neg} & \textcolor{green!40!black}{pos} & 	 horrendous well well nice nice
\\
Top-$k$ gradient 	& \textcolor{orange!70!black}{neg} & \textcolor{green!40!black}{pos} & 	 well horrendous this well very
\\
Top-$k$ softmax 	& \textcolor{orange!70!black}{neg} & \textcolor{green!40!black}{pos} & 	 horrendous well Oh well nice
\\
Top-$k$ 1.5-entmax 	& \textcolor{orange!70!black}{neg} & \textcolor{orange!70!black}{neg} & 	 horrendous Oh surprised had agree
\\
Top-$k$ sparsemax 	& \textcolor{orange!70!black}{neg} & \textcolor{green!40!black}{pos} & 	 horrendous smart nice Oh had
\\
Select. entmax15 	& \textcolor{orange!70!black}{neg} & \textcolor{green!40!black}{pos} & 	 horrendous Oh surprised had agree nice smart others ever well ever but most nice movie proud like wonder . way few without . find but It making well actually be everyday
\\
Select. sparsemax 	& \textcolor{orange!70!black}{neg} & \textcolor{green!40!black}{pos} & 	 horrendous smart nice Oh had ever ever few . wonder nice
\\
Bernoulli 	& \textcolor{orange!70!black}{neg} & \textcolor{green!40!black}{pos} & 	 most horrendous bunch extras mostly scenes few This movie old movie everyday lot nice extras wrestler wrestler nice guy
\\
HardKuma 	& \textcolor{orange!70!black}{neg} & \textcolor{orange!70!black}{neg} & 	 horrendous bunch few confused wonder Oh guy very smart
\\
Joint $E$ and $L$ 	& \textcolor{orange!70!black}{neg} & \textcolor{orange!70!black}{neg} & 	 horrendous without
\\

\toprule
\\
\multicolumn{4}{p{16cm}}{\textbf{(\textcolor{green!40!black}{positive})} Having read some of the other comments here I was expecting something truly awful but was pleasantly surprised . REALITY CHECK : The original series wasn ' t that good . I think some people remember it with more affection than it deserved but apart from the car chases and Daisy Duke ' s legs the scripts were weak and poorly acted . The Duke boys were too intelligent and posh for backwood hicks , the shrunken Boss Hog was too cretinous to be evil and Rosco was just hyper throughout every screen moment . It ' s amazing the series actually lasted as long as it did because it ran out of story lines during the first series . $<$ br $>$ $<$ br $>$ Back to the movie . If you watch this film in it ' s own right , not as a direct comparison to however you remember the TV series , then it ' s not bad at all . The real star is of course the General Lee . The car chases and stunts are excellent and that ' s really what D . O . H . is all about . Johnny Knoxville is his usual eccentric self and along with Seann William Scott as Cousin Bo the pair make this film really funny in a hilarious Dumb - And - Dumber sort of way the TV series never achieved . The lovely Jessica Simpson is a natch as Miss Daisy , Burt Reynolds makes a much improved Boss Hog and M . C . Gainey makes a believably nasty Rosco P . Coltrane , the way he always should have been . $<$ br $>$ $<$ br $>$ If you don ' t like slapstick humour and crazy car stunts then you wouldn ' t be watching this film anyway because you should know what to expect . Otherwise if you want an entertaining car - action movie with a few good laughs that ' s not too taxing on the brain then go see this enjoyable romp with an open mind .
} \\

\midrule

\sc Explainer & $y_C$ & $y_L$  &  \sc Explanation \\

\midrule

Erasure         	& \textcolor{green!40!black}{pos} & \textcolor{green!40!black}{pos} & 	 horrendous well well nice nice
\\
Top-$k$ gradient 	& \textcolor{green!40!black}{pos} & \textcolor{orange!70!black}{neg} & 	 well horrendous this well very
\\
Top-$k$ softmax 	& \textcolor{green!40!black}{pos} & \textcolor{green!40!black}{pos} & 	 horrendous well Oh well nice
\\
Top-$k$ 1.5-entmax 	& \textcolor{green!40!black}{pos} & \textcolor{green!40!black}{pos} & 	 horrendous Oh surprised had agree
\\
Top-$k$ sparsemax 	& \textcolor{green!40!black}{pos} & \textcolor{orange!70!black}{neg} & 	 horrendous smart nice Oh had
\\
Select. entmax15 	& \textcolor{green!40!black}{pos} & \textcolor{green!40!black}{pos} & 	 horrendous Oh surprised had agree nice smart others ever well ever but most nice movie proud like wonder . way few without . find but It making well actually be everyday
\\
Select. sparsemax 	& \textcolor{green!40!black}{pos} & \textcolor{green!40!black}{pos} & 	 horrendous smart nice Oh had ever ever few . wonder nice
\\
Bernoulli 	        & \textcolor{green!40!black}{pos} & \textcolor{green!40!black}{pos} & 	 most horrendous bunch extras mostly scenes few This movie old movie everyday lot nice extras wrestler wrestler nice guy
\\
HardKuma 	        & \textcolor{orange!70!black}{neg} & \textcolor{orange!70!black}{neg} & 	 horrendous bunch few confused wonder Oh guy very smart
\\
Joint $E$ and $L$ 	& \textcolor{green!40!black}{pos} & \textcolor{green!40!black}{pos} & 	 horrendous without
\\

\bottomrule
\end{tabular}
% }
\end{center}
\caption{(continuation) Examples of extracted explanations for IMDB.} 
\label{table:examples_explanations_imdb_csrs2}
\end{table*}

\begin{table*}[!htb]
\scriptsize
\begin{center}
% \resizebox{\textwidth}{!}{%
\begin{tabular}{lllp{10cm}}

\toprule

\textbf{(\textcolor{green!40!black}{entailment})} \\

\multicolumn{4}{p{15cm}}{Premise:  Children and adults swim in large pool with red staircase .} \\

\multicolumn{4}{p{15cm}}{Hypothesis:  A group of people are swimming .}
\\

\midrule

\sc Explainer & $y_C$ & $y_L$  &  \sc Explanation \\

\midrule

Erasure 	         & \textcolor{green!40!black}{ent} & \textcolor{green!40!black}{ent} & 	 swim pool staircase adults
\\
Top-$k$ gradient 	 & \textcolor{green!40!black}{ent} & \textcolor{orange!70!black}{con} & 	 adults pool swim large
\\
Top-$k$ softmax 	 & \textcolor{green!40!black}{ent} & \textcolor{green!40!black}{ent} & 	 swim pool large staircase
\\
Top-$k$ 1.5-entmax 	 & \textcolor{green!40!black}{ent} & \textcolor{green!40!black}{ent} & 	 swim pool large staircase
\\
Top-$k$ sparsemax 	 & \textcolor{green!40!black}{ent} & \textcolor{green!40!black}{ent} & 	 swim pool large adults
\\
Select. entmax15 	 & \textcolor{green!40!black}{ent} & \textcolor{green!40!black}{ent} & 	 swim pool large staircase adults Children in and with
\\
Select. sparsemax 	 & \textcolor{green!40!black}{ent} & \textcolor{green!40!black}{ent} & 	 swim pool large adults in
\\
Bernoulli 	         & \textcolor{green!40!black}{ent} & \textcolor{green!40!black}{ent} & 	 Children and adults swim in large pool with red staircase .
\\
HardKuma 	         & \textcolor{green!40!black}{ent} & \textcolor{orange!70!black}{con} & 	 swim large pool staircase
\\
Joint $E$ and $L$ 	 & \textcolor{green!40!black}{ent} & \textcolor{orange!70!black}{con} & 	 pool swim staircase
\\

\midrule
\\

\textbf{(\textcolor{orange!70!black}{contradiction})} \\

\multicolumn{4}{p{15cm}}{Premise:  A group of Asian children are gathered around in a circle listening to an older male in a white shirt .} \\

\multicolumn{4}{p{15cm}}{Hypothesis:  A man is wearing a black shirt .}
\\

\midrule

\sc Explainer & $y_C$ & $y_L$  &  \sc Explanation \\

\midrule

Erasure 	        & \textcolor{orange!70!black}{con} & \textcolor{green!40!black}{ent} & 	 Asian white male children
\\
Top-$k$ gradient 	 & \textcolor{orange!70!black}{con} & \textcolor{green!40!black}{ent} & 	 circle children gathered to
\\
Top-$k$ softmax 	 & \textcolor{orange!70!black}{con} & \textcolor{green!40!black}{ent} & 	 Asian white male children
\\
Top-$k$ 1.5-entmax 	 & \textcolor{orange!70!black}{con} & \textcolor{orange!70!black}{con} & 	 white older a male
\\
Top-$k$ sparsemax 	 & \textcolor{orange!70!black}{con} & \textcolor{orange!70!black}{con} & 	 a male shirt Asian
\\
Select. entmax15 	 & \textcolor{orange!70!black}{con} & \textcolor{orange!70!black}{con} & 	 white older a male Asian listening circle shirt of children around gathered a an group in in . A to are
\\
Select. sparsemax 	 & \textcolor{orange!70!black}{con} & \textcolor{orange!70!black}{con} & 	 a male shirt Asian . an
\\
Bernoulli 	        & \textcolor{green!40!black}{ent} & \textcolor{green!40!black}{ent} & 	 A group of Asian children are gathered around in a circle listening to an older male in a white shirt .
\\
HardKuma 	        & \textcolor{orange!70!black}{con} & \textcolor{green!40!black}{ent} & 	 group Asian male white shirt
\\
Joint $E$ and $L$ 	 & \textcolor{orange!70!black}{con} & \textcolor{orange!70!black}{con} & 	 male group white
\\

\midrule
\\

\textbf{(\textcolor{orange!70!black}{contradiction})} \\

\multicolumn{4}{p{15cm}}{Premise:  A woman is pushing her bike with a baby carriage in front .} \\

\multicolumn{4}{p{15cm}}{Hypothesis:  A woman is pushing groceries in a cart .}
\\

\midrule

\sc Explainer & $y_C$ & $y_L$  &  \sc Explanation \\

\midrule

Erasure 	         & \textcolor{orange!70!black}{con} & \textcolor{orange!70!black}{con} & 	 baby woman bike pushing
\\
Top-$k$ gradient 	 & \textcolor{orange!70!black}{con} & \textcolor{orange!70!black}{con} & 	 carriage bike her with
\\
Top-$k$ softmax 	 & \textcolor{orange!70!black}{con} & \textcolor{blue!70!black}{neu} & 	 baby woman carriage pushing
\\
Top-$k$ 1.5-entmax 	 & \textcolor{orange!70!black}{con} & \textcolor{orange!70!black}{con} & 	 carriage woman her baby
\\
Top-$k$ sparsemax 	 & \textcolor{green!40!black}{ent} & \textcolor{orange!70!black}{con} & 	 baby carriage woman front
\\
Select. entmax15 	 & \textcolor{orange!70!black}{con} & \textcolor{orange!70!black}{con} & 	 carriage woman her baby pushing front is A . a bike with
\\
Select. sparsemax 	 & \textcolor{green!40!black}{ent} & \textcolor{green!40!black}{ent} & 	 baby carriage woman front pushing is
\\
Bernoulli 	         & \textcolor{orange!70!black}{con} & \textcolor{orange!70!black}{con} & 	 A woman is pushing her bike with a baby carriage in front .
\\
HardKuma 	         & \textcolor{orange!70!black}{con} & \textcolor{orange!70!black}{con} & 	 woman pushing bike carriage
\\
Joint $E$ and $L$ 	 & \textcolor{orange!70!black}{con} & \textcolor{orange!70!black}{con} & 	 woman baby
\\

\midrule
\\

\textbf{(\textcolor{blue!70!black}{neutral})} \\

\multicolumn{4}{p{15cm}}{Premise:   A woman in a gray shirt working on papers at her desk .} \\

\multicolumn{4}{p{15cm}}{Hypothesis:  Lady working in her desk tensely to completed the task}
\\

\midrule

\sc Explainer & $y_C$ & $y_L$  &  \sc Explanation \\

\midrule

Erasure 	         & \textcolor{blue!70!black}{neu} & \textcolor{blue!70!black}{neu} & 	 desk papers woman .
\\
Top-$k$ gradient 	 & \textcolor{blue!70!black}{neu} & \textcolor{blue!70!black}{neu} & 	 desk on shirt at
\\
Top-$k$ softmax 	 & \textcolor{blue!70!black}{neu} & \textcolor{blue!70!black}{neu} & 	 desk papers woman .
\\
Top-$k$ 1.5-entmax 	 & \textcolor{blue!70!black}{neu} & \textcolor{blue!70!black}{neu} & 	 desk papers working woman
\\
Top-$k$ sparsemax 	 & \textcolor{blue!70!black}{neu} & \textcolor{blue!70!black}{neu} & 	 desk papers woman working
\\
Select. entmax15 	 & \textcolor{blue!70!black}{neu} & \textcolor{green!40!black}{ent} & 	 desk papers working woman . on shirt her at in a
\\
Select. sparsemax 	 & \textcolor{blue!70!black}{neu} & \textcolor{green!40!black}{ent} & 	 desk papers woman working her A
\\
Bernoulli 	         & \textcolor{blue!70!black}{neu} & \textcolor{blue!70!black}{neu} & 	 A woman in a gray shirt working on papers at her desk .
\\
HardKuma 	         & \textcolor{blue!70!black}{neu} & \textcolor{blue!70!black}{neu} &   woman working papers at desk
\\
Joint $E$ and $L$ 	 & \textcolor{blue!70!black}{neu} & \textcolor{blue!70!black}{neu} & 	 working desk woman papers
\\

\midrule
\\

\textbf{(\textcolor{blue!70!black}{neutral})} \\

\multicolumn{4}{p{15cm}}{Premise:  A brown dog with a blue muzzle is running on green grass .} \\

\multicolumn{4}{p{15cm}}{Hypothesis:  A mean dog is wearing a muzzle to keep it from attacking cats}
\\

\midrule

\sc Explainer & $y_C$ & $y_L$  &  \sc Explanation \\

\midrule

Erasure 	         & \textcolor{blue!70!black}{neu} & \textcolor{blue!70!black}{neu} & 	 dog brown running muzzle
\\
Top-$k$ gradient 	 & \textcolor{blue!70!black}{neu} & \textcolor{blue!70!black}{neu} & 	 with brown on green
\\
Top-$k$ softmax 	 & \textcolor{blue!70!black}{neu} & \textcolor{blue!70!black}{neu} & 	 dog brown running blue
\\
Top-$k$ 1.5-entmax 	 & \textcolor{orange!70!black}{con} & \textcolor{orange!70!black}{con} & 	 dog blue brown muzzle
\\
Top-$k$ sparsemax 	 & \textcolor{blue!70!black}{neu} & \textcolor{blue!70!black}{neu} & 	 dog muzzle with is
\\
Select. entmax15 	 & \textcolor{orange!70!black}{con} & \textcolor{blue!70!black}{neu} & 	 dog blue brown muzzle running is . A grass green with on a
\\
Select. sparsemax 	 & \textcolor{blue!70!black}{neu} & \textcolor{blue!70!black}{neu} & 	 dog muzzle with is A a running on brown
\\
Bernoulli 	         & \textcolor{blue!70!black}{neu} & \textcolor{orange!70!black}{con} & 	 A brown dog with a blue muzzle is running on green grass .
\\
HardKuma 	         & \textcolor{blue!70!black}{neu} & \textcolor{blue!70!black}{neu} & 	 dog muzzle running
\\
Joint $E$ and $L$ 	 & \textcolor{blue!70!black}{neu} & \textcolor{blue!70!black}{neu} & 	 dog running muzzle
\\

\bottomrule
\end{tabular}
% }
\end{center}
\caption{Examples of extracted explanations for SNLI.} 
\label{table:examples_explanations_snli_csrs}
\end{table*}

\begin{table*}[!htb]
\scriptsize
\begin{center}
% \resizebox{\textwidth}{!}{%
\begin{tabular}{lllp{10cm}}

\toprule

\textbf{(\textcolor{orange!70!black}{contradiction})} \\

\multicolumn{4}{p{15cm}}{Premise:  A man sits at a table in a room .} \\

\multicolumn{4}{p{15cm}}{Hypothesis:  A woman sits .}
\\

\midrule

\sc Explainer & $y_C$ & $y_L$  &  \sc Explanation \\

\midrule

Erasure 	 & \textcolor{orange!70!black}{con} & \textcolor{green!40!black}{ent} & 	 sits table . at
\\
Top-$k$ gradient 	 & \textcolor{orange!70!black}{con} & \textcolor{green!40!black}{ent} & 	 . sits table A
\\
Top-$k$ softmax 	 & \textcolor{orange!70!black}{con} & \textcolor{green!40!black}{ent} & 	 sits table . room
\\
Top-$k$ 1.5-entmax 	 & \textcolor{orange!70!black}{con} & \textcolor{green!40!black}{ent} & 	 table . sits man
\\
Top-$k$ sparsemax 	 & \textcolor{orange!70!black}{con} & \textcolor{green!40!black}{ent} & 	 man sits A at
\\
Select. entmax15 	 & \textcolor{orange!70!black}{con} & \textcolor{green!40!black}{ent} & 	 table . sits man A room a a at in
\\
Select. sparsemax 	 & \textcolor{orange!70!black}{con} & \textcolor{green!40!black}{ent} & 	 man sits A at in a a
\\
Bernoulli 	 & \textcolor{orange!70!black}{con} & \textcolor{green!40!black}{ent} & 	 A man sits at a table in a room .
\\
HardKuma 	 & \textcolor{orange!70!black}{con} & \textcolor{orange!70!black}{con} & 	 man sits at
\\
Joint $E$ and $L$ 	 & \textcolor{orange!70!black}{con} & \textcolor{orange!70!black}{con} & 	 man
\\
Human Highlights 	 & \textcolor{orange!70!black}{con} & \textcolor{green!40!black}{ent} & 	 man
\\

\midrule
\\

\textbf{(\textcolor{green!40!black}{entailment})} \\

\multicolumn{4}{p{15cm}}{Premise:  Elderly woman climbing up the stairs .} \\

\multicolumn{4}{p{15cm}}{Hypothesis:  The old lady was walking up the stairs .}
\\

\midrule

\sc Explainer & $y_C$ & $y_L$  &  \sc Explanation \\

\midrule

Erasure 	 & \textcolor{green!40!black}{ent} & \textcolor{green!40!black}{ent} & 	 stairs woman Elderly climbing
\\
Top-$k$ gradient 	 & \textcolor{green!40!black}{ent} & \textcolor{orange!70!black}{con} & 	 Elderly stairs . the
\\
Top-$k$ softmax 	 & \textcolor{green!40!black}{ent} & \textcolor{orange!70!black}{con} & 	 stairs woman Elderly climbing
\\
Top-$k$ 1.5-entmax 	 & \textcolor{green!40!black}{ent} & \textcolor{orange!70!black}{con} & 	 stairs Elderly woman climbing
\\
Top-$k$ sparsemax 	 & \textcolor{green!40!black}{ent} & \textcolor{orange!70!black}{con} & 	 stairs Elderly woman climbing
\\
Select. entmax15 	 & \textcolor{green!40!black}{ent} & \textcolor{orange!70!black}{con} & 	 stairs Elderly woman climbing up . the
\\
Select. sparsemax 	 & \textcolor{green!40!black}{ent} & \textcolor{orange!70!black}{con} & 	 stairs Elderly woman climbing the
\\
Bernoulli 	 & \textcolor{orange!70!black}{con} & \textcolor{orange!70!black}{con} & 	 Elderly woman climbing up the stairs .
\\
HardKuma 	 & \textcolor{green!40!black}{ent} & \textcolor{orange!70!black}{con} & 	 Elderly woman climbing up stairs
\\
Joint $E$ and $L$ 	 & \textcolor{green!40!black}{ent} & \textcolor{green!40!black}{ent} & 	 stairs Elderly climbing woman
\\
Human Highlights 	 & \textcolor{green!40!black}{ent} & \textcolor{orange!70!black}{con} & 	 Elderly woman climbing
\\

\midrule
\\

\textbf{(\textcolor{green!40!black}{entailment})} \\

\multicolumn{4}{p{15cm}}{Premise:  A woman with a blond ponytail and a white hat is riding a white horse , inside a fence with a horned cow .} \\

\multicolumn{4}{p{15cm}}{Hypothesis:  The woman is riding a horse .}
\\

\midrule

\sc Explainer & $y_C$ & $y_L$  &  \sc Explanation \\

\midrule

Erasure 	 & \textcolor{green!40!black}{ent} & \textcolor{orange!70!black}{con} & 	 horse riding . fence
\\
Top-$k$ gradient 	 & \textcolor{green!40!black}{ent} & \textcolor{green!40!black}{ent} & 	 cow horse fence riding
\\
Top-$k$ softmax 	 & \textcolor{green!40!black}{ent} & \textcolor{green!40!black}{ent} & 	 horse riding fence cow
\\
Top-$k$ 1.5-entmax 	 & \textcolor{green!40!black}{ent} & \textcolor{orange!70!black}{con} & 	 horse riding woman a
\\
Top-$k$ sparsemax 	 & \textcolor{green!40!black}{ent} & \textcolor{orange!70!black}{con} & 	 horse riding a is
\\
Select. entmax15 	 & \textcolor{green!40!black}{ent} & \textcolor{orange!70!black}{con} & 	 horse riding woman a cow fence horned a is ponytail , a with inside blond A . hat
\\
Select. sparsemax 	 & \textcolor{green!40!black}{ent} & \textcolor{orange!70!black}{con} & 	 horse riding a is with A ,
\\
Bernoulli 	 & \textcolor{green!40!black}{ent} & \textcolor{orange!70!black}{con} & 	 A woman with a blond ponytail and a white hat is riding a white horse , inside a fence with a horned cow .
\\
HardKuma 	 & \textcolor{green!40!black}{ent} & \textcolor{orange!70!black}{con} & 	 woman ponytail riding horse inside horned cow
\\
Joint $E$ and $L$ 	 & \textcolor{green!40!black}{ent} & \textcolor{green!40!black}{ent} & 	 cow horse fence inside
\\
Human Highlights 	 & \textcolor{green!40!black}{ent} & \textcolor{green!40!black}{ent} & 	 woman blond horse fence horned cow
\\

\midrule
\\

\textbf{(\textcolor{orange!70!black}{contradiction})} \\

\multicolumn{4}{p{15cm}}{Premise:   A woman in a black coat eats dinner while her dog looks on .} \\

\multicolumn{4}{p{15cm}}{Hypothesis:  A woman is wearing a blue coat .}
\\

\midrule

\sc Explainer & $y_C$ & $y_L$  &  \sc Explanation \\

\midrule

Erasure 	 & \textcolor{orange!70!black}{con} & \textcolor{green!40!black}{ent} & 	 coat black woman dog
\\
Top-$k$ gradient 	 & \textcolor{orange!70!black}{con} & \textcolor{green!40!black}{ent} & 	 dog eats black looks
\\
Top-$k$ softmax 	 & \textcolor{orange!70!black}{con} & \textcolor{green!40!black}{ent} & 	 coat black woman dinner
\\
Top-$k$ 1.5-entmax 	 & \textcolor{orange!70!black}{con} & \textcolor{orange!70!black}{con} & 	 black coat woman dog
\\
Top-$k$ sparsemax 	 & \textcolor{orange!70!black}{con} & \textcolor{orange!70!black}{con} & 	 black a woman A
\\
Select. entmax15 	 & \textcolor{orange!70!black}{con} & \textcolor{orange!70!black}{con} & 	 black coat woman dog a looks in . dinner eats her A on while
\\
Select. sparsemax 	 & \textcolor{orange!70!black}{con} & \textcolor{green!40!black}{ent} & 	 black a woman A coat in her
\\
Bernoulli 	 & \textcolor{orange!70!black}{con} & \textcolor{green!40!black}{ent} & 	 A woman in a black coat eats dinner while her dog looks on .
\\
HardKuma 	 & \textcolor{orange!70!black}{con} & \textcolor{green!40!black}{ent} & 	 woman black coat
\\
Joint $E$ and $L$ 	 & \textcolor{orange!70!black}{con} & \textcolor{orange!70!black}{con} & 	 woman black
\\
Human Highlights 	 & \textcolor{orange!70!black}{con} & \textcolor{orange!70!black}{con} & 	 black
\\

\bottomrule
\end{tabular}
% }
\end{center}
\caption{(continuation) Examples of extracted explanations for SNLI.} 
\label{table:examples_explanations_snli_csrs2}
\end{table*}

\end{document}